# How Well Can Large Language Models Impersonate Individual Authors? A Comparative Evaluation

Harshit Gupta, Shunichi Ishihara

Speech and Language Laboratory, Australian National University, Australia,
harshit.gupta@anu.edu.au, shunichi.ishihara@anu.edu.au

**Abstract:** This study investigates the authorship impersonation capacity of six widely used large language models (LLMs): Gemini 3 Pro, Claude Opus, GPT 5.2, GPT 5.2 Pro, DeepSeek 3.2, and Grok 4.1 Fast. The evaluation was conducted using short product-review texts selected to ensure high within-author consistency in individual writing style. To reflect plausible misuse scenarios in which wrongdoers are likely to favour straightforward and technically undemanding methods, two types of in-context prompt were designed: a simple prompt with minimal instruction and a complex prompt incorporating explicit stylometric guidance for mimicry. Authorship mimicry was evaluated using a cosine-distance-based distance-to-training metric, supplemented by visual analyses of embedding distributions. The results indicate that authorship impersonation remained challenging for most LLMs, with Gemini 3 Pro showing the strongest apparent mimicry performance. However, subsequent overlap analysis revealed that Gemini 3 Pro extensively copied and reused sentences and phrases from the training texts, at levels far exceeding those observed among human authors. Follow-up factor analyses showed that training-set consistency was the most robust predictor of mimicry success across models, whereas author distinctiveness did not generalise as a reliable predictor. Global embedding visualisations further revealed a consistent macro-level pattern: for most models, generated texts occupied regions of the embedding space that were largely separated from the corresponding human training clusters. This indicates limited convergence towards target-author style and suggests the presence of model-specific lexical and phraseological preferences. This interpretation was further supported by supervised source-attribution experiments distinguishing between human-written texts and texts generated by the six LLMs. The findings highlight the importance of distinguishing genuine stylistic imitation from direct textual reuse and have implications for the forensic analysis of textual evidence involving potential LLM-enabled impersonation.

## 1 Introduction

Writing style impersonation has a long and well-documented history, ranging from medieval political forgeries (Britannica Editors 2016) and literary hoaxes (Bonner and Noble 2025) to contemporary criminal cases involving digital communication. In modern forensic contexts, the Danielle Jones and Jenny Nicholl cases are particularly well known (Grant 2008). These cases are directly relevant to forensic linguistics because the perpetrators attempted to evade detection by impersonating the victims through SMS messages, deliberately reproducing subtle stylistic cues in order to mislead criminal investigations.

The forensic analysis of textual evidence has therefore long recognised that writers may deliberately disguise, alter, or manipulate their writing style. Until recently, however, writing style impersonation was necessarily a human activity. The emergence and rapid development of generative artificial intelligence (AI), particularly large language models (LLMs), has introduced a new and potentially more scalable form of this problem. LLMs are increasingly capable of producing natural language that has been reported to be difficult for humans to distinguish from human-written text (Jakesch et al. 2023; Kreps et al. 2022; Porter and Machery 2024). If misused, these systems may facilitate impersonation by enabling malicious actors to imitate the writing styles of trusted individuals (Alotaibi et al. 2024; Shah et al. 2023). This risk is amplified by the widespread availability of affordable, user-friendly AI chatbots and by the large volume of personal text data that can be readily obtained online. Together, these developments make it increasingly easy to generate deceptive messages at scale for identity impersonation, fraud, and identity theft (Blauth et al. 2022; Broklyn et al. 2024). Reflecting these concerns, Europol warned in its 2023 report that LLMs may be misused for fraud, social engineering, and phishing, specifically noting their capacity to reproduce language patterns and impersonate the styles of particular individuals or groups (European Union Agency for Law Enforcement Cooperation 2023).

Under these circumstances, it is important to understand the actual potential of LLMs for writing style impersonation, as well as the nature of machine-generated texts in comparison with human-written texts. Many earlier studies examining the ability of LLMs to impersonate or mimic distinctive writing styles have focused on well-known targets, including canonical authors (Jhamtani et al. 2017; Kirilloff et al. 2025; Mikros 2025), public figures (Herbold et al. 2024), as well as politicians, celebrities, and fictional characters (Li et al. 2023). In such settings, models may be able to draw on the comparatively large and publicly available textual footprints associated with these targets, making their styles, personas, or discourse patterns easier to learn and reproduce than those of ordinary individuals (Wang et al. 2025). Much of this work has also been conducted within the broader paradigm of text style transfer, often as benchmarking or evaluation studies (Jin et al. 2022; Mukherjee et al. 2024; Toshevska and Gievska 2021; Yang et al. 2018). More recent research has extended this line of inquiry to ordinary individuals whose available textual data may be sparse, approaching the problem through the lens of authorship attribution and authorship verification (Alperin et al. 2025; Bhandarkar et al. 2024; Chen and Moscholios 2024; Ishihara 2026; Jemama and Kumar 2025b; Thompson and Ishihara 2025; Wang et al. 2025; Zeng and Nini 2026).

From a forensic perspective, it is essential not only to assess whether authorship mimicry occurs, but also to examine the nature of the resulting texts; for example, why one LLM appears to mimic an authorial style more successfully than another (Ballantyne et al. 2024). Accordingly, the present study investigates the characteristics of several widely used LLMs in terms of their capacity for authorship mimicry, while systematically comparing their mimicry performance under a common evaluation framework. Because mimicry performance may vary across target authors, the study also examines whether some authors are more readily impersonated than others, and identifies the properties that distinguish easier-to-mimic authors from those whose styles are more resistant to imitation. In addition, the study analyses how machine-generated texts differ from human-written texts across the tested LLMs. Finally, as emphasised by Thompson and Ishihara (2025) and Ishihara (2026), the impersonation capacity of LLMs should be assessed periodically in light of their rapidly advancing capabilities. The present study contributes to this ongoing monitoring effort.

Various approaches have been proposed to enable LLMs to learn and reproduce the writing styles of individual authors. One approach is author-specific fine-tuning, in which a model is separately adapted to each target author by updating its parameters so that its outputs better reflect that author's distinctive stylistic features. This method has been shown to be effective for personalised text generation (Tan et al. 2024; Zhang et al. 2024). An alternative approach follows a one-shot or few-shot prompting paradigm (Brown et al. 2020). In this setting, an LLM is given a small number of sample texts from a target author as in-context evidence, allowing it to infer and emulate aspects of the author's writing style without parameter updating. Given that wrongdoers are likely to favour straightforward and cost-effective methods requiring limited technical expertise (Isola 2024), the present study adopts this latter approach. Specifically, it evaluates the ability of LLMs to mimic personal writing styles using in-context prompting with one-shot author-specific evidence.

The remainder of this paper is organised as follows. §2 reviews related work on LLM-based stylistic imitation, highlighting key methodologies and remaining research gaps. §3 describes the study methodology, including the selection of authors and texts, the partitioning of author-specific texts, the selection of the embedding model used in subsequent analyses, and the establishment of key reference values. §4 details the tested LLMs, text-generation procedures, prompt designs, and assessment metrics. §5 presents the results, focusing on model-specific differences in mimicry performance and author-specific factors associated with susceptibility to imitation. §6 summarises the main findings, and §7 discusses their implications, outlines the limitations of the study, and identifies directions for future work.

## 2 Related Work

A closely related line of work to writing-style impersonation—and, more broadly, writing-style imitation—is text style transfer (TST) (Toshevska and Gievska 2021). TST aims to modify specific stylistic attributes of a text while preserving its underlying propositional content (Jin et al. 2022; Mukherjee et al. 2024). It has been applied to a wide range of tasks, including formality transfer (Madaan et al. 2020), sentiment manipulation, text simplification, detoxification (Dale et al. 2021), learning support (Kostopolus 2025), authorship obfuscation (Kacmarcik and Gamon 2006; Shokri et al. 2025), and author-style imitation (de Boer et al. 2023; Lingard 2023; Mikros 2025). Two of these applications—authorship obfuscation and author-style imitation—are particularly relevant to the present study and have legitimate uses. Authorship obfuscation can help protect vulnerable writers from re-identification (Wang 2023), while author-style imitation may support genre acquisition or personalised writing adapted to particular audiences or interlocutors for inclusiveness (Jin et al. 2022; Wang et al. 2024).

Although these technologies have been developed for legitimate and socially beneficial purposes, they may also be exploited by malicious actors. For example, an offender could use an LLM to compose a fraudulent message that mimics the writing style of a trusted person, such as a manager or family member, thereby increasing the likelihood that the recipient will regard the message as genuine. Similarly, such systems could be used to produce disinformation or fake news in the apparent style of a target individual or victim, who may then be publicly criticised or held responsible for content they did not write. Some studies further demonstrate that individual authors can be imitated by LLMs in such ways that human readers fail to detect (Herbold et al. 2024; Shi et al. 2025).

Although LLM-generated writing style imitation may be difficult for human readers to detect, its effectiveness against authorship verification systems remains contested. Zeng and Nini (2026) report that texts generated by GPT-4o under several prompting conditions, with the aim of imitating the writing style of a target individual, failed to evade established authorship verification systems. Their findings suggest that current systems may remain robust against relatively simple, prompt-based impersonation attempts. By contrast, Alperin et al. (2025) show that modern authorship verification systems can be substantially perturbed by more strategic LLM-based attacks, reporting maximum attack success rates of 92% for authorship obfuscation and 78% for authorship impersonation using open-source language models. Bhandarkar et al. (2024) examined to what extend 12 off-the-shelf LLMs can perform author style emulation through prompted inputs with directed instructions such as stylometric information. They reported clear differences in style-emulation performance across the tested models, but found that, overall, the models fell short of achieving effective author style emulation. They also emphasised the advantage of user-specific instructions for improving generation quality. Chen and Moscholios (2024) also emphasise the importance of prompt design with conversational AIs, in particular reporting effectiveness of Tree-of-Thoughts prompting for improved authorship imitation ability. Likewise, Jemama and Kumar (2025a) also report the importance of prompting strategies has a substantial influence on the successfully style imitation. Taken together, these findings confirm the important role of prompt designs and strategies in successful authorship impersonation.

Although the importance of prompt engineering has been widely acknowledged in prior research (Sahoo et al. 2024; Vatsal and Dubey 2024), individual writing style is often subtle, diffuse, and implicitly encoded across texts (Grieve 2023). As a result, eliciting such stylistic characteristics may require carefully crafted prompts. Designing prompts capable of drawing out nuanced personal style, however, typically demands specialised technical expertise and detailed stylistic awareness (Kestemont 2014). In misuse contexts, malicious actors are unlikely to invest such effort, instead favouring low-cost and readily accessible techniques, such as straightforward in-context learning, particularly when the anticipated reward does not justify more labour-intensive prompt optimisation. Under such conditions, Wang et al. (2025) investigated the mimicry capacity of four conventional LLMs using a limited number of per-author text samples across four textual domains. Their findings indicate that LLMs struggled to reproduce nuanced individual styles, particularly in informal and stylistically heterogeneous domains such as forums and blogs, while achieving comparatively better performance in more structured domains, including news articles and emails.

Overall, previous studies suggest that the success of AI-based authorship impersonation depends strongly on the experimental setting, the model used, and the sophistication of the prompting or generation strategy. Under low-effort conditions, such as simple in-context prompting with only a small amount of author-specific training data, LLMs appear to have difficulty reproducing nuanced individual writing styles and may fail to evade established authorship verification systems. Nevertheless, existing findings also indicate that this difficulty is not uniform: some models perform better than others, and some authors appear to be more readily mimicked than others. To the best of our knowledge, however, it remains unclear why certain LLMs achieve stronger authorship impersonation performance than others, and which textual or stylistic properties make some authors easier to imitate. The present study addresses these questions by evaluating the authorship impersonation capacity of recent LLMs available as of January 2026, while examining both model-specific differences in mimicry performance and author-specific factors associated with susceptibility to imitation.

## 3 Methodology

This section describes the procedure used to select the authors and their human-written texts for use as in-context evidence in LLM-based authorship mimicry, including the author-consistency analysis applied during selection. It also explains how the selected texts for each author were partitioned into separate groups to support multiple text generations for the same target author. Before this selection process, the embedding models were benchmarked to identify the model that most effectively separated same-author (SA) and different-author (DA) pairs of human-written texts. The selected embedding model was then used for author-consistency scoring, author and text selection, mimicry evaluation, and subsequent analyses. This benchmarking procedure also established key reference values, including the threshold that best separates SA and DA human-written text pairs and the corresponding distance distributions.[1] These values serve as reference points in the analyses that follow.

[1] The codes used in this study is available at xxx.

### 3.1 Problem Statement

The authorship impersonation task is formulated as follows. Given a human-written text $T_A$ by author $A$, an LLM is instructed to generate a text $T_G$ that imitates the writing style represented in $T_A$. Task success is then evaluated using an authorship metric that assesses whether $T_A$ and $T_G$ are sufficiently similar to be judged as having been written by the same author.

For this evaluation, a decision threshold $\tau$ is defined using cosine distance. The threshold is selected to maximise separation between SA pairs of human-written texts and DA pairs of human-written texts. Impersonation is considered successful if the cosine distance between $T_A$ and $T_G$ is lower than $\tau$. In other words, when $T_G$ falls within the SA region relative to $T_A$, the LLM is judged to have successfully imitated the writing style of author $A$.

### 3.2 Base Database Description

The Amazon Authorship Verification Corpus (AAVC)[2] (Halvani et al. 2017) was used as the basis for selecting the text sets analysed in this study (see Figure 2). The AAVC comprises Amazon product reviews written by 3,227 reviewers (authors), totalling 21,347 reviews. As shown in Figure 1A, the number of reviews per author varies, with most authors contributing between five and seven reviews, and a minimum and maximum of two and 16 reviews, respectively. As illustrated in Figure 1B, the reviews span 17 product and service categories. Each review is standardised to 4 KB (approximately 700 words). Within each author's set, review categories are non-duplicative; that is, no author contributes multiple reviews within the same category.

[2] http://bit.ly/1OjFRhJ

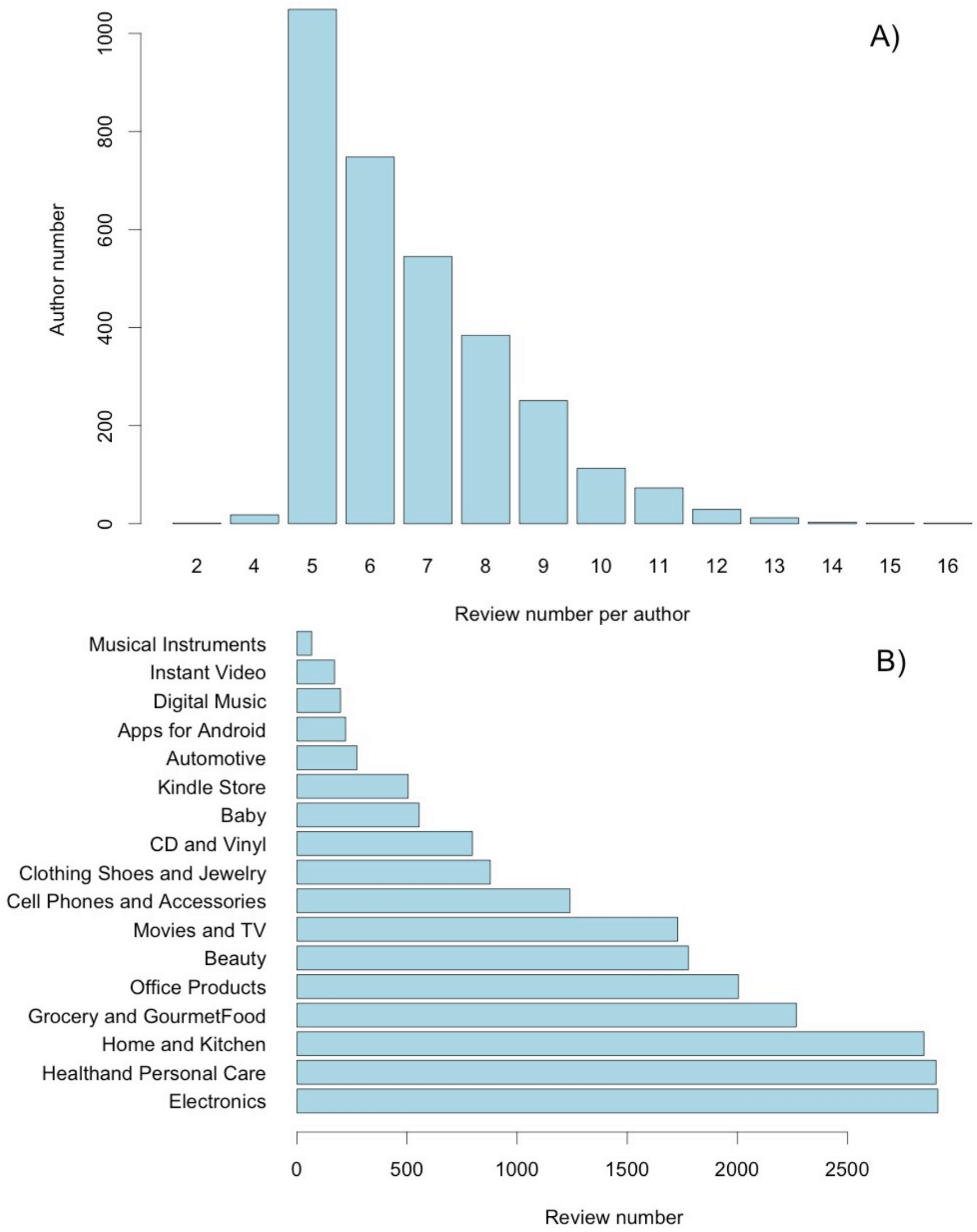

Figure 1: Basic statistics of the AAVC corpus. (A) Histogram of reviews per author. (B) List of the 17 product categories (topics) with corresponding review counts.

From the full pool of 3,227 authors, those with at least six reviews were retained to ensure sufficient topic coverage for controlled within-author partitioning. This filtering step yielded 2,144 eligible authors, which formed the base pool for embedding-model benchmarking (§3.3) and for subsequent identification of authors with high stylistic consistency (§3.4). As detailed in §3.4, 100 authors whose six selected reviews exhibited strong within-author consistency were ultimately chosen to provide the training evidence for the LLM mimicry experiments. Figure 2 summarises the multi-stage selection procedure leading to the final set of 100 authors.

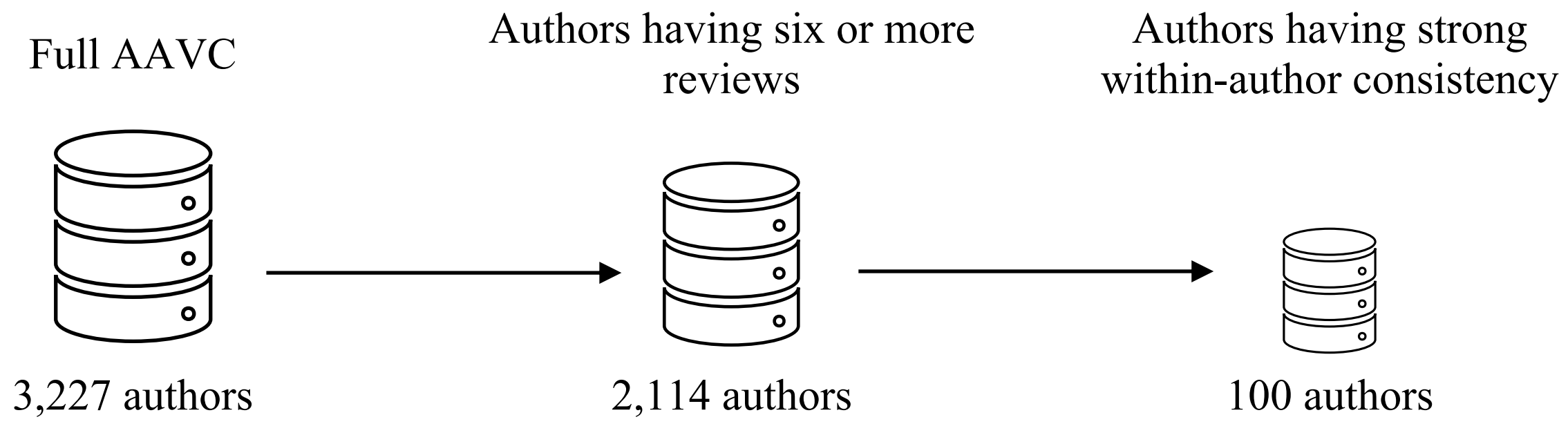

Figure 2: Steps leading to the final set of 100 authors.

Note that the base pool of 2,144 authors and their reviews was used to benchmark the embedding models in §3.3.

### *3.3 Embedding-model Benchmarking and Authorship-consistency Analysis for Author Selection*

The following six embedding models for author style representation were benchmarked to identify a representation that best separate SA pairs and DA pairs of texts.

Model 1. rrivera1849/LUAR-CRUD[3]
Model 2. rrivera1849/LUAR-MUD[4]
Model 3. gabrielloiseau/LUAR-CRUD-sentence-transformers[5]
Model 4. gabrielloiseau/LUAR-MUD-sentence-transformers[6]
Model 5. AnnaWegmann/Style-Embedding[7]
Model 6. AIDA-UPM/star[8]

Models 1 and 2 are the original LUAR (Learning Universal Authorship Representations) models (Rivera-Soto et al. 2021), trained on a subsample of the Pushshift Reddit dataset (CRUD) and on the Reddit Million User Dataset (MUD), respectively. Models 3 and 4 are sentence-transformer variants of those LUAR models. Model 5 is a Sentence-Transformers model designed for writing-style embeddings (Wegmann et al. 2022), and Model 6 is the original STAR (Style Transformer for Authorship Representations) model (Huertas-Tato et al. 2024). Detailed model specifications are provided in the footnote links.

Benchmarking followed the ROC–AUC (receiver operating characteristic–area under the curve) protocol (Fawcett 2006), as specified below, and used cosine distance as the primary similarity measure.

1. Two reviews were randomly selected for each author in the base pool (2,114 authors).
2. A total of 2,144 SA review pairs were constructed.
3. A total of 2,144 DA pairs were sampled; for each DA pair, one of the four possible cross-comparisons was selected at random, yielding 2,144 DA review pairs in total.
4. Cosine distance was computed for all SA and DA review pairs.
5. ROC curves and associated summary statistics were generated.
6. The best-performing embedding model was selected on the basis of AUC.

[3] https://huggingface.co/rrivera1849/LUAR-CRUD
[4] https://huggingface.co/rrivera1849/LUAR-MUD
[5] https://huggingface.co/gabrielloiseau/LUAR-CRUD-sentence-transformers
[6] https://huggingface.co/gabrielloiseau/LUAR-MUD-sentence-transformers
[7] https://huggingface.co/AnnaWegmann/Style-Embedding
[8] https://huggingface.co/AIDA-UPM/star

The benchmarking protocol has been repeated ten times. Table 1 reports the summary statistics from the benchmarking used to identify the highest-performing embedding model.

Table 1: Summary statistics of the ten repeats of the benchmarking protocol (SD = standard deviation; dist. = cosine distance). The highest AUC and Cohen's *d* values are shown in Bold. The mean threshold $\tau$ of Model 2 is underlined.

| Rank | Model | Mean AUC | SD AUC | Mean Threshold $\tau$ | Mean Cohen's *d* | Mean SA dist. | Mean DA dist. |
|---|---|---|---|---|---|---|---|
| 1 | Model 2 | **0.9614** | 0.0024 | <u>0.2648</u> | **2.4603** | 0.1967 | 0.3713 |
| 2 | Model 1 | 0.9546 | 0.0016 | 0.1998 | 2.3182 | 0.1484 | 0.2824 |
| 3 | Model 6 | 0.9356 | 0.0032 | 0.1960 | 2.0902 | 0.1379 | 0.2933 |
| 4 | Model 4 | 0.8898 | 0.0041 | 0.3961 | 1.7162 | 0.3303 | 0.4775 |
| 5 | Model 3 | 0.8897 | 0.0026 | 0.3998 | 1.7175 | 0.3328 | 0.4817 |
| 6 | Model 5 | 0.8763 | 0.0073 | 0.1212 | 1.2600 | 0.0780 | 0.3296 |

Models 1, 2, and 6 demonstrate strong performance (AUC > 0.9) relative to the other models (AUC < 0.9). Models 1 and 2, and likewise Models 3 and 4, share the same underlying architectures but differ in their training data. Accordingly, each pair achieves comparable performance (Models 1 and 2: AUC = 0.9546 and 0.9614, respectively; Models 3 and 4: AUC = 0.8897 and 0.8898, respectively). The SDs of AUC are very small (< 0.008), indicating consistent performance across repeats. On the basis of ROC–AUC, Model 2 (LUAR_MUD)—the original LUAR model (Rivera-Soto et al. 2021) trained on a subsample of the CRUD—was selected as the primary embedding model for downstream analyses. Model 2 also yields the largest Cohen's *d* (2.4603), further supporting this choice. That is, this selection of the model fixed the embedding space used for author consistency scoring, document selection, mimicry evaluation, and subsequent analyses.

In the benchmarking exercise, the mean threshold $\tau$ (i.e., the optimal distance) for Model 2 was 0.2648 (Table 1). The corresponding mean SA and DA cosine distances were 0.1967 and 0.3713, respectively. These values will be used as reference points in subsequent analyses.

As explained earlier, the benchmarking protocol was repeated ten times (Repeats 1–10). Figure 3 shows the cosine-distance distributions for the SA and DA pairs for Repeats 4 and 7, which yielded the lowest (AUC = 0.9571) and highest (AUC = 0.9666) AUC values, respectively, across the ten repeats. Figure 3 therefore illustrates the extent of overlap between the SA and DA distributions in the worst- and best-performing repeats.

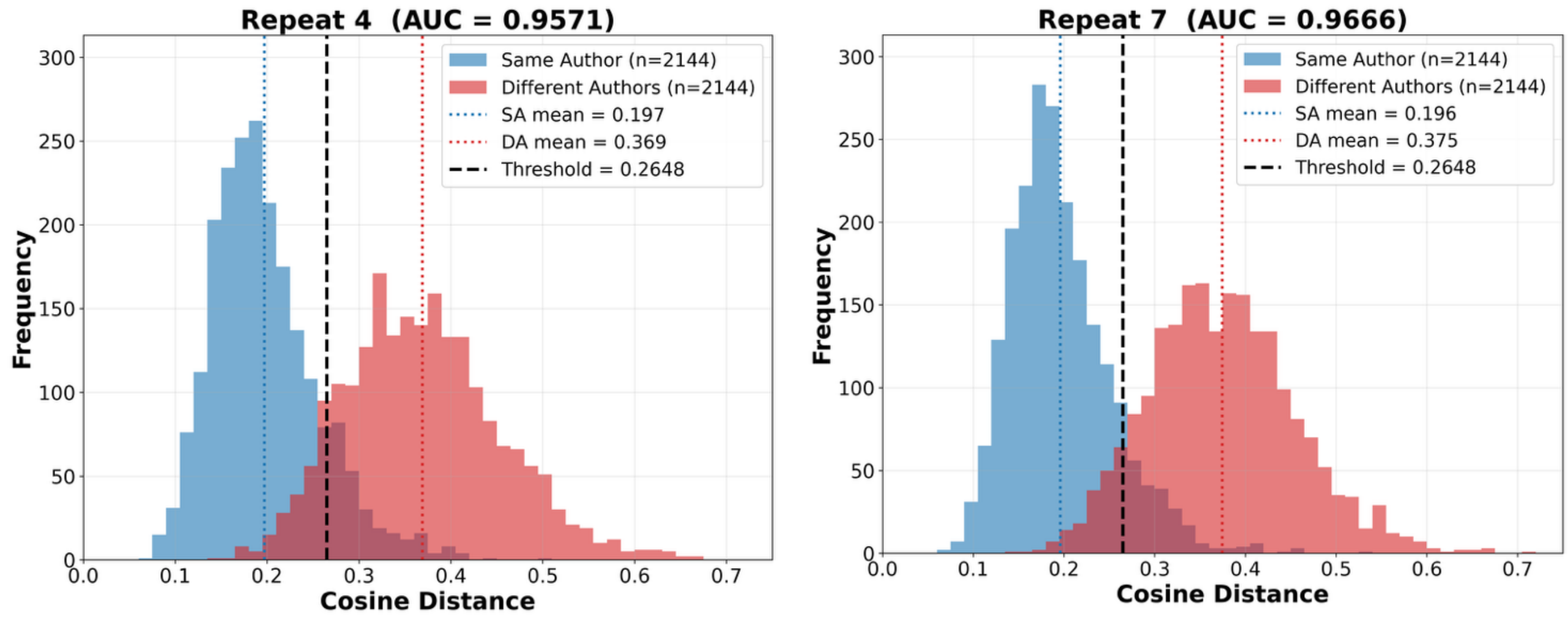

Figure 3: Cosine-distance distributions for SA (blue) and DA (red) pairs in Repeat 4 (left) and Repeat 7 (right).

Figure 3 shows that the SA and DA cosine-distance distributions overlap to some extent. This overlap implies that, for some authors, pairs of reviews written by the same individual can nevertheless appear dissimilar in the embedding space and may be indistinguishable from DA pairs. Such authors and reviews should be excluded from the training data used to model individual writing styles with LLMs. Accordingly, subsequent analyses should focus on authors whose reviews exhibit sufficient within-author consistency to support reliable characterisation of their distinctive writing patterns.

### *3.4 Consistency Analysis and Selection of Authors and Reviews for Training*

A consistency analysis was conducted to prioritise authors whose reviews exhibit stable stylistic representations. Here, stability was defined as low within-author dispersion in the embedding space. This procedure produced a ranked list of authors, from which the 100 most consistent were selected for the text-generation experiments (see Figure 2). Specifically, from the base pool of 2,144 authors and their reviews used for benchmarking in §3.3, the top 100 authors and six representative reviews per author were selected using the following steps:

1. For each author ($A$), compute the centroid of that author's review embeddings.
2. Identify the six reviews ($AR_1 \sim AR_6$) closest to the centroid.
3. Compute the mean cosine distance of all possible pairs ($15 = \binom{6}{2}$) of these six reviews.
4. Rank and qualify the 2,144 authors over their full review set: an author is retained only if it has at least six reviews and its maximum pairwise SA cosine distance falls below the 0.2648 threshold, and the 100 authors with the smallest mean pairwise distance across all their reviews are selected. For each selected author, the six representative reviews from Steps 1–2 form the training set whose 15-pair statistics (Step 3) are summarised in Table 2.

Step 3 is important because it ensures that the selected six reviews are appropriately consistent and can be attributed to the same author, as reflected in the mean cosine distance relative to the benchmark values in Table 1.

A summary of basic statistics for the mean cosine distances computed in Step 3 for the selected top 100 authors is provided in Table 2.

Table 2: Summary statistics of the top 100 authors' mean cosine-distances.

| Overall Mean | Max | Min | SD |
|---|---|---|---|
| 0.1313 | 0.1497 | 0.0925 | 0.0118 |

Table 2 shows that, across the 100 authors, the overall mean distance was 0.1313, with author-level mean distances ranging from 0.0925 to 0.1497. These values are substantially lower than the benchmark overall mean SA distance of 0.1967 reported in Table 1 and illustrated in Figure 3. This indicates that the six selected reviews for each author exhibit strong within-author stylistic consistency. Given that the benchmark threshold separating SA and DA pairs was 0.2648, review pairs drawn from each of the 100 selected authors would be expected to fall well within the SA range.

Figure 4 shows a two-dimensional $t$-SNE ($t$-Distributed Stochastic Neighbour Embedding) (van der Maaten and Hinton 2008), of author-level centroids. Specifically, the centroids of the six selected reviews for each of the 100 target authors are plotted alongside the centroids of the remaining AAVC authors. All embeddings were generated using Model 2, LUAR_MUD.

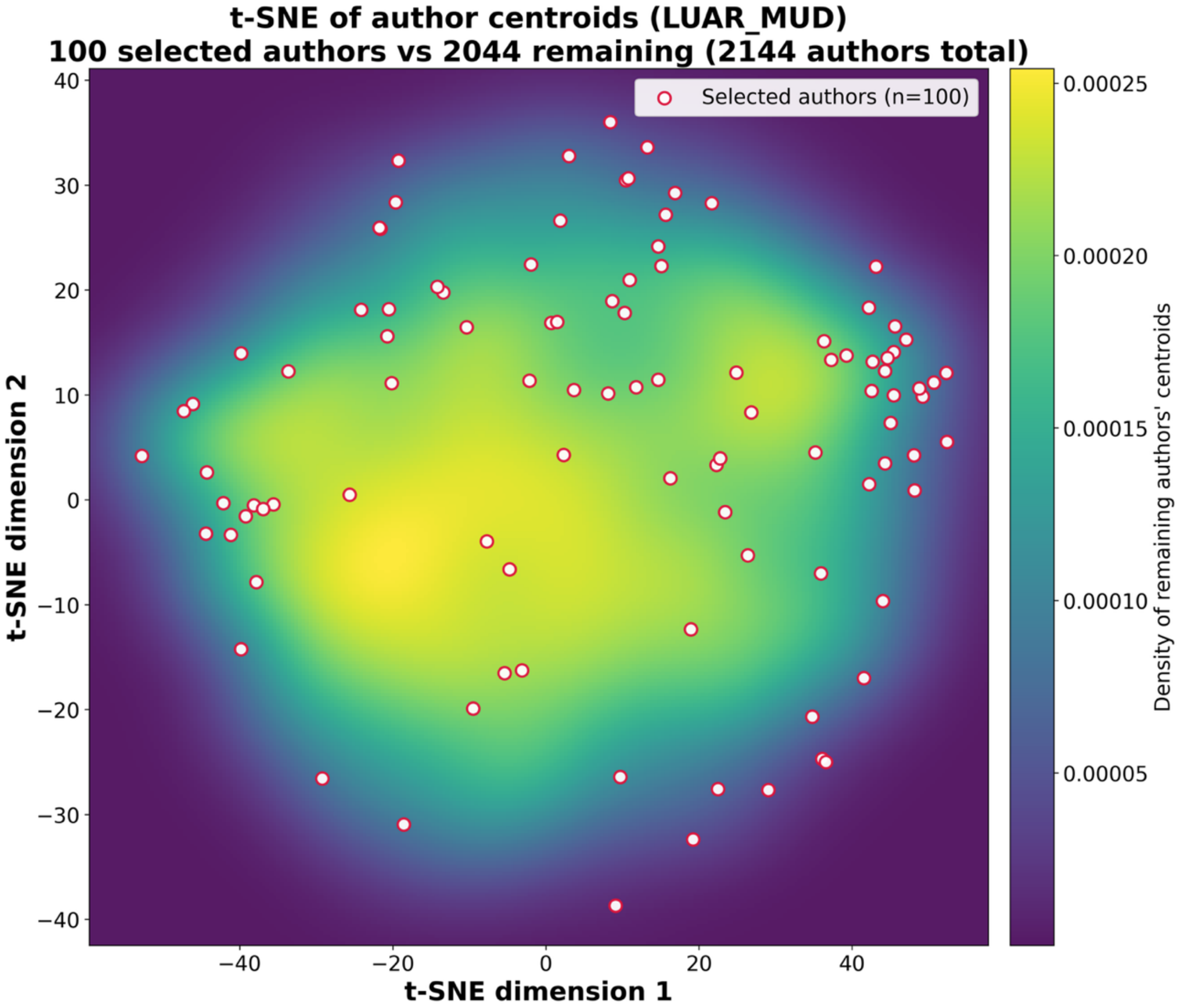

Figure 4: Two-dimensional *t*-SNE projection of author centroids. Circles represent the centroids of the six-review sets for the 100 selected authors, while the heatmap represents the distribution of centroids for the remaining authors in the corpus.

The centroids of the 100 selected authors are distributed across the projection space, indicating substantial stylistic diversity and broad coverage of the range observed in the full corpus.

### *3.4.1 Grouping of Reviews for Training*

The training texts for the authorship mimicry task were therefore established using the six most internally consistent reviews for each author. For each author, these six reviews (e.g., $R_1$ to $R_6$) were randomly partitioned into two groups (G1 and G2) of three, as illustrated in Figure 5 using the second author ($A_2$) as an example. Each three-review group served as the in-context evidence for a separate generation context, thereby improving robustness by avoiding reliance on a single fixed set of exemplars. Consequently, two texts were machine generated for each author (e.g., $MA_2R_1$ and $MA_2R_2$).

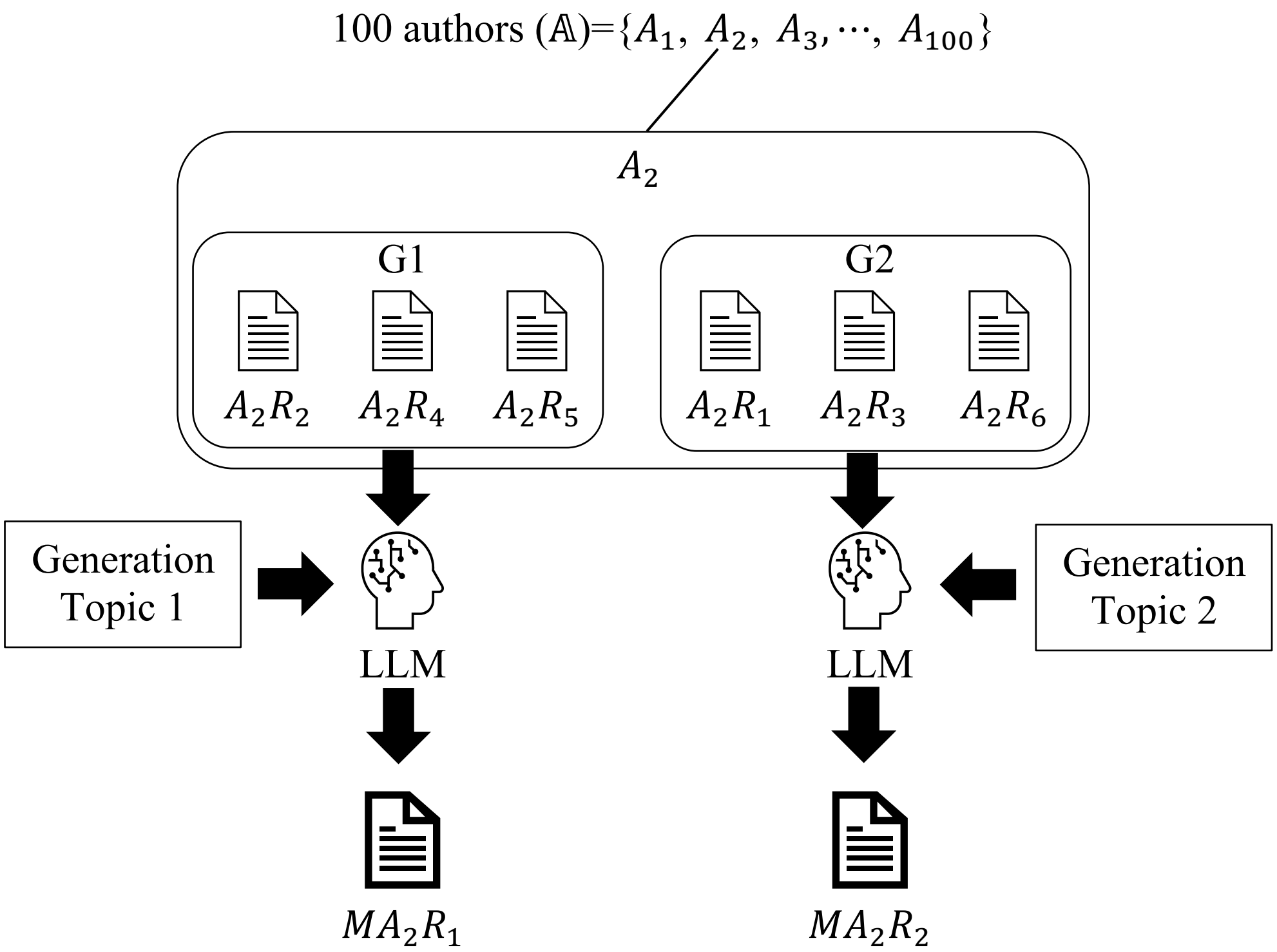

Figure 5: Illustration of the grouping of reviews used as in-context evidence for text generation. G1 = the first group of reviews; G2 = the second group of reviews. $A_2R_2$, for example, denotes the second review written by $A_2$. $MA_2R_1$, for example, denotes a machine-generated review based on the first group of reviews written by $A_2$.

## 4 Large Language Models for Text Generation and Mimicry Performance Metric

Synthetic review generation was executed across six LLMs listed in Table 3[9] to compare mimicry capability under the same experimental design.

Table 3: Tested LLMs.

| Provider | Models | Full Keys |
|---|---|---|
| OpenAI | GPT 5.2 | gpt-5.2-2025-12-11 |
| OpenAI | GPT 5.2 Pro | gpt-5.2-pro-2025-12-11 |
| Google | Gemini 3 Pro | gemini-3-pro-preview |
| Anthropic | Claude Opus | claude-opus-4-5-20251101 |
| xAI | Grok 4.1 Fast | grok-4-1-fast-reasoning |
| DeepSeek | DeepSeek 3.2 | deepseek-reasoner |

A prompt-generation script was implemented to standardise how training evidence and instructions were presented to each LLM. The script assembled author-specific examples, selected held-out generation topics, and produced both simple and complex prompt variants to enable controlled comparisons. The actual prompts used to generate the reviews are provided in Appendix A.

[9] Several other LLMs (Llama, Copilot, and Perplexity) were not included in the study due to technical and practical constraints at the time the study was conducted. Llama was excluded because API access remains restricted under a waitlist-based rollout and, in practice, is primarily oriented towards US-based developers. Copilot was excluded because its Chat API is relatively new, remains in preview, and is designed chiefly for interactive, grounded enterprise use rather than controlled offline text generation. Perplexity was excluded because policy and legal gating led to repeated refusals, preventing reliable and consistent completion of the mimicry prompts.

Generation topics were selected to differ from the training-topic set in each generation context, such that the task assessed style transfer rather than topic memorisation. As noted in §3.2, each author's reviews span multiple product or service categories with no overlap, across 17 topics in total. Following the example in Figure 3, the reviews in G1 ($A_2R_2, A_2R_4, MA_2R_5$) and those in G2 ($A_2R_1, A_2R_3, MA_2R_6$) each cover distinct topics. Generation Topic 1 was therefore sampled at random from the topics not represented in the G1 reviews, and Generation Topic 2 was sampled at random from the topics not represented in the G2 reviews, with the additional constraint that Generation Topic 2 differed from Generation Topic 1.

As noted above, two prompts were prepared for each training group: a simple prompt and a complex prompt. The full prompt contexts are provided in Appendix A. The simple prompt provided only minimal instructions, whereas the complex prompt included explicit stylistic guidance and additional constraints, including a strong anti-copying instruction. The simple prompt was designed to approximate a plausible AI misuse scenario involving impersonation, in which the perpetrator is likely to be a layperson providing only minimal instructions (Isola 2024).

Under both prompt conditions, each model was instructed to generate a 600–900-word review, corresponding to approximately 4 KB of text. The temperature was set to 0.7 for all LLMs. This design produced four generated texts per author, allowing within-author comparisons across both prompt variants and training contexts. Table 4 summarises the number of generated review texts at each level of aggregation.

Table 4: Summary of the number of generated reviews.

| Level | Computation | Total |
|---|---|---|
| Per author | 2 training groups × 2 prompt variants | 4 generations |
| Per LLM | 100 authors × 4 generations | 400 generations |
| Across LLMs | 6 LLMs × 400 generations | 2,400 generations |

Although the LLMs generally followed the specified review-length instruction, they occasionally produced longer outputs. To minimise length-related confounds in embedding behaviour and subsequent similarity calculations, all generated texts were normalised to a target length of approximately 4 KB while preserving sentence boundaries. This is consistent with the AAVC procedure described in §3.2, in which review texts are standardised to 4 KB. The normalised generated texts were then embedded using the selected authorship embedding model (LUAR_MUD). To further improve robustness, a split-half embedding procedure was applied, followed by averaging of the resulting embeddings.

Authorship mimicry performance was assessed using a "distance-to-training" metric, defined as the cosine distance between the generated text and the corresponding training texts. This procedure is illustrated in Figure 6, which reuses the example presented in Figure 5.

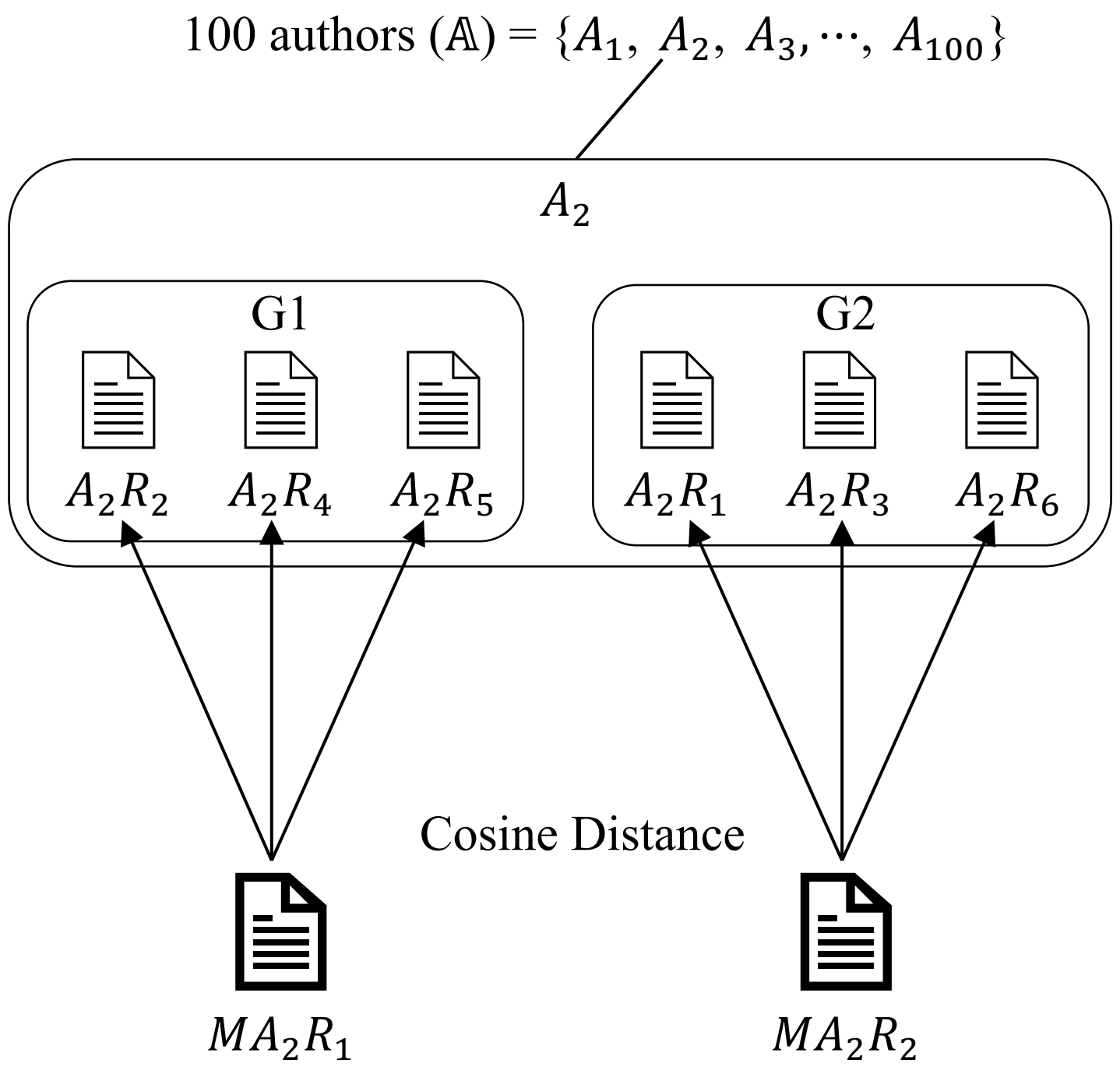

Figure 6: Illustration of distance-to-training metric using the example given in Figure 5.

# 5 Results

## *5.1 Mimicry Performance*

Model performance was ranked using the "distance-to-training" metric, with lower values indicating closer alignment between each generated review and the author's training reviews. For each LLM, an overall metric score was also computed as the mean across the two prompt types (simple and complex), as reported in Table 5.

Table 5: Mean "distance-to-training" metric values. Δ denotes the difference between the two prompts (Simple − Complex); positive values indicate better performance under the complex prompt than under the simple prompt.

| Rank | LLM | Mean Simple | Mean Complex | Overall Mean | Δ |
|---|---|---|---|---|---|
| 1 | Gemini 3 Pro | 0.3052 | 0.2955 | 0.3003 | 0.0097 |
| 2 | Claude Opus | 0.3845 | 0.3695 | 0.3770 | 0.0149 |
| 3 | GPT 5.2 Pro | 0.4661 | 0.4505 | 0.4350 | 0.0155 |
| 4 | GPT 5.2 | 0.4484 | 0.4495 | 0.4490 | –0.0011 |
| 5 | DeepSeek 3.2 | 0.4874 | 0.4797 | 0.4835 | 0.0076 |
| 6 | Grok 4.1 Fast | 0.4470 | 0.5323 | 0.4896 | –0.0853 |

Table 5 shows that Gemini 3 Pro achieved the lowest overall "distance-to-training" score (0.3003), whereas Grok 4.1 fast produced the highest (0.4896). For context, the benchmarking results in §2.2 using LUAR_MUD yielded mean SA and DA cosine distances of 0.1967 and 0.3713, respectively (Table 1). If authorship mimicry were fully successful, the mean distance between an author's training reviews and the corresponding machine-generated reviews would be expected to approach the within-author baseline (i.e., the SA mean). Although Gemini 3 Pro performs best among the evaluated LLMs, its mean distance (0.3003) remains substantially higher than the benchmark SA mean (0.1967). This suggests that, while Gemini 3 Pro exhibits the greatest mimicry potential in the present set-up, it does not achieve the level of similarity observed between genuine SA texts. Turning to the remaining LLMs, their overall mean distances are close to, or exceed, the benchmark DA mean (0.3713). This indicates limited mimicry: the generated reviews are likely to be judged as DA texts relative to their corresponding training reviews. Overall, while the tested LLMs differ in their capacity for authorship impersonation, mimicry appears generally limited, with Gemini 3 Pro representing the strongest—though still imperfect—case.

Prompt-type effects were evaluated by comparing the "distance-to-training" scores obtained under the simple and complex prompts within each model, using $\Delta$ as the simple–complex difference. For Gemini 3 Pro ($\Delta$ = 0.0097), Claude Opus ($\Delta$ = 0.0149), GPT 5.2 Pro ($\Delta$ = 0.0155), and DeepSeek-V3.2 ($\Delta$ = 0.0097), the complex prompt yielded slightly lower distances, consistent with improved mimicry. By contrast, GPT 5.2 ($\Delta$ = –0.0011) and Grok 4.1 Fast ($\Delta$ = –0.0853) performed better under the simple prompt, suggesting that the more elaborate instruction format did not translate into improved mimicry for these models under the present experimental conditions. Overall, the effect of prompt type appears to be model-specific, but generally small and, for some models, negligible.

Table 6 shows the extent to which the generated reviews are assessed as having been written by the same author as the training reviews. For this test, the two groups of reviews were compared using the SA/DA threshold of 0.2648 (refer to §3.3). If the distance between the generated and training reviews is below 0.2648, they are assessed as having the same author, and the impersonation is considered successful.

Table 6: Writing style mimicry success rate (%) at the SA/DA threshold of 0.2648—percentage of the 100 reviews generated per prompt whose distance to the training data falls below 0.2648 (i.e. assessed as SA, indicating successful impersonation).

| Rank | LLM | Simple | Complex | Overall |
|---|---|---|---|---|
| 1 | Gemini 3 Pro | 29.0 | 36.0 | 32.5 |
| 2 | Claude Opus | 2.0 | 3.0 | 2.5 |
| 3 | GPT 5.2 Pro | 0.0 | 0.0 | 0.0 |
| 4 | GPT 5.2 | 0.0 | 0.0 | 0.0 |
| 5 | DeepSeek 3.2 | 0.0 | 0.0 | 0.0 |
| 6 | Grok 4.1 Fast | 3.0 | 0.0 | 1.5 |

As with the observations from Table 5, Table 6 also indicates the limited mimicry capability of the tested LLMs. Gemini 3 Pro demonstrates the strongest performance, with 29% and 36% of the generated reviews produced using the simple and complex prompts, respectively, successfully mimicking the target authors' writing styles. In contrast, the other models achieve mimicry success rates of only 0–3%.

The limited mimicry potential of the tested LLMs, as summarised in Table 5 and Table 6, suggests that human-written and machine-generated reviews remain distinguishable and exhibit different writing-style characteristics (Zaitsu and Jin 2023; Zaitsu et al. 2024, 2025). To investigate this issue more closely, a global multi-author visualisation was constructed. This visualisation was designed to display all authors simultaneously, together with both their training documents and generated texts, thereby enabling inspection of corpus-wide clustering patterns, identification of outliers, and visual comparison of the proximity of simple- and complex-prompt generations to their corresponding training clusters.

For the visualisation, first of all, the embeddings were $L_2$-normalised to unit length prior to dimensionality reduction to align with cosine geometry, and then *t*-SNE and UMAP (Uniform Manifold Approximation and Projection) (McInnes et al. 2018) were applied in two dimensions. As representative visualisations, *t*-SNE plots for GPT 5.2, Claude Opus, and Gemini 3 Pro are presented as Figure 7, Figure 8 and Figure 9, respectively in the main text, as the overall trends are consistent across the two plot types.

As can be seen in Figure 7, which a *t*-SNE plot for GPT 5.2, a consistent macro-structure was observed across most LLMs. For GPT 5.2, GPT 5.2 Pro, DeepSeek V3.2, and Grok 4.1 Fast, the generated reviews occupied regions that were largely separated from the human training reviews. This separation suggested that, at a corpus level, the generated texts often did not lie in the same local neighbourhoods as the author training documents, and as also suggested by the distance-based assessment given in Table 5, therefore, it can be seen that style convergence to the target authors was limited in the embedding space used for evaluation.

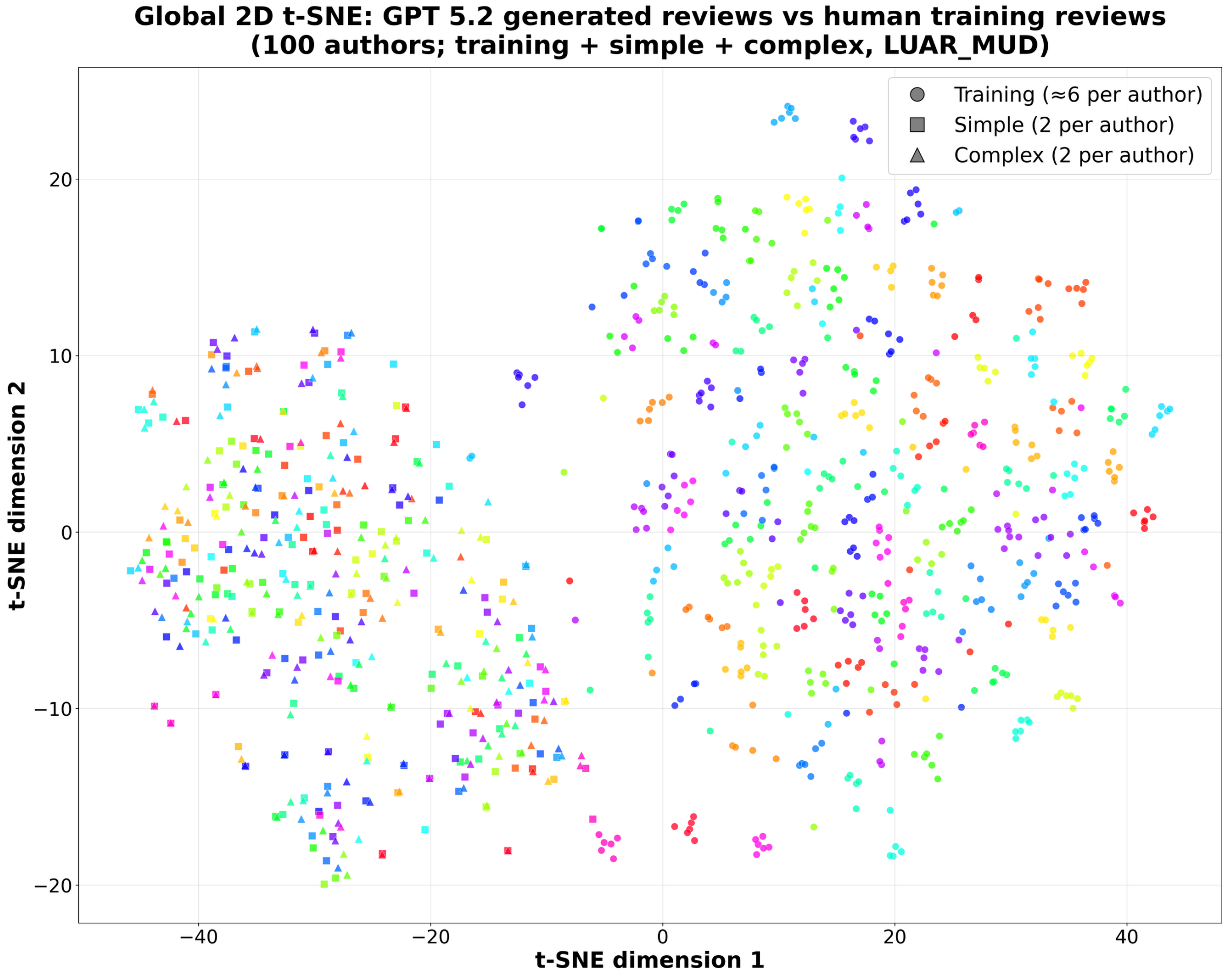

Figure 7: Global two-dimensional *t*-SNE plot of GPT 5.2 generated reviews under the simple and complex prompts (two reviews per author per prompt), together with the training reviews (six per author).

Figure 8 presents a *t*-SNE plot for Claude Opus, showing somewhat more heterogeneous behaviour than that observed for GPT-5.2, GPT-5.2 Pro, DeepSeek-V3.2, and Grok 4.1 Fast. In particular, the Claude-generated texts form two clusters, one substantially larger than the other. For a small subset of authors, the generated points interleave with, or lie close to, the corresponding training clusters, indicating partial stylistic alignment. This pattern is consistent with Claude Opus ranking second in Table 5. However, a substantial proportion of Claude-generated reviews remain separated from the training documents, indicating that mimicry was not achieved consistently across authors.

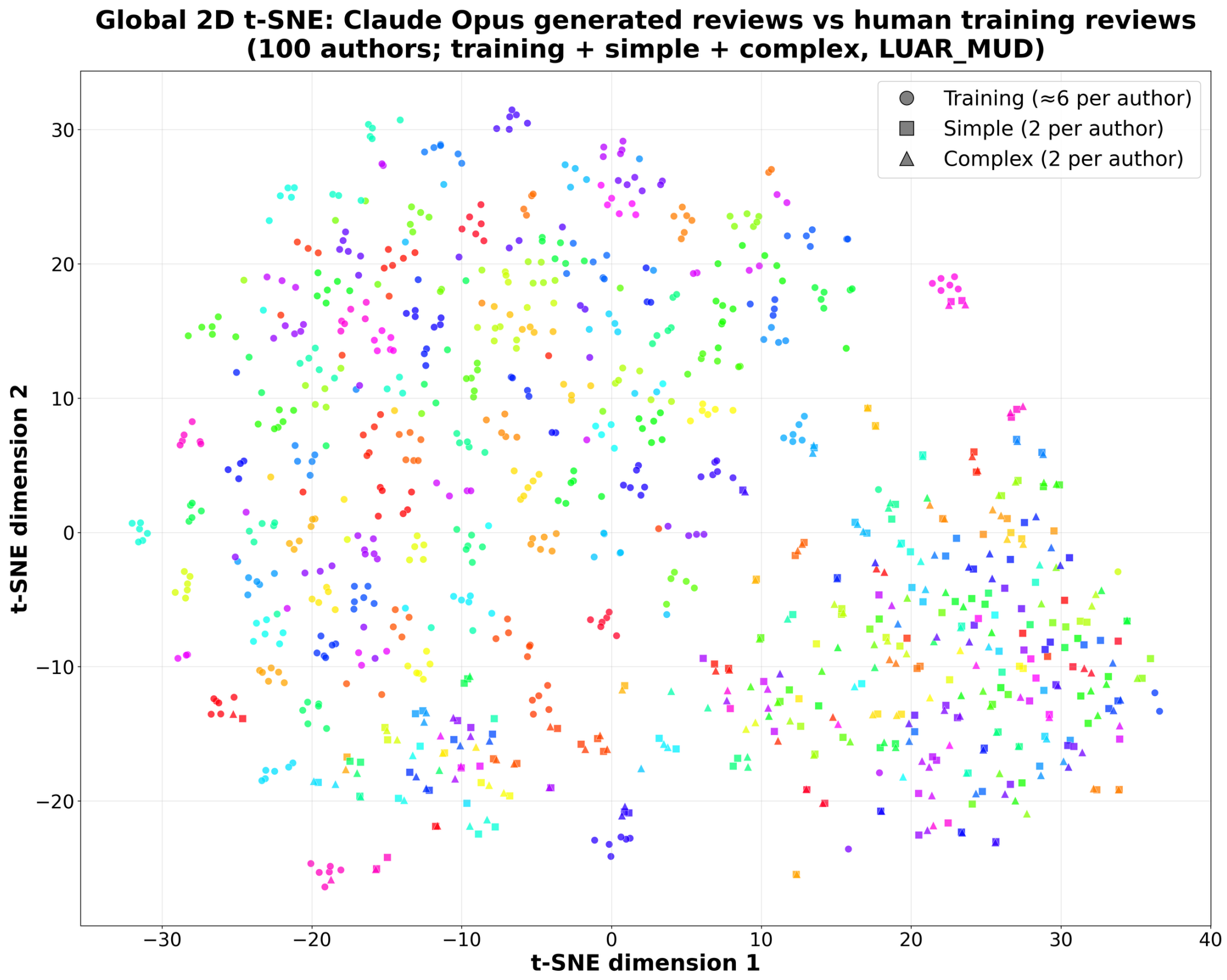

Figure 8: Global two-dimensional *t*-SNE plot of Claude Opus generated reviews under the simple and complex prompts (two reviews per author per prompt), together with the training reviews (six per author).

Figure 9 presents a *t*-SNE plot for Gemini 3 Pro and shows the strongest alignment between generated and training documents. The generated points are frequently located within, or immediately adjacent to, the corresponding training clusters, although an apparent tight cluster of Gemini-generated reviews is also observed. Relative to the other LLMs, this greater spatial proximity is consistent with Gemini 3 Pro achieving the best performance on the distance-to-training metric reported in Table 5.

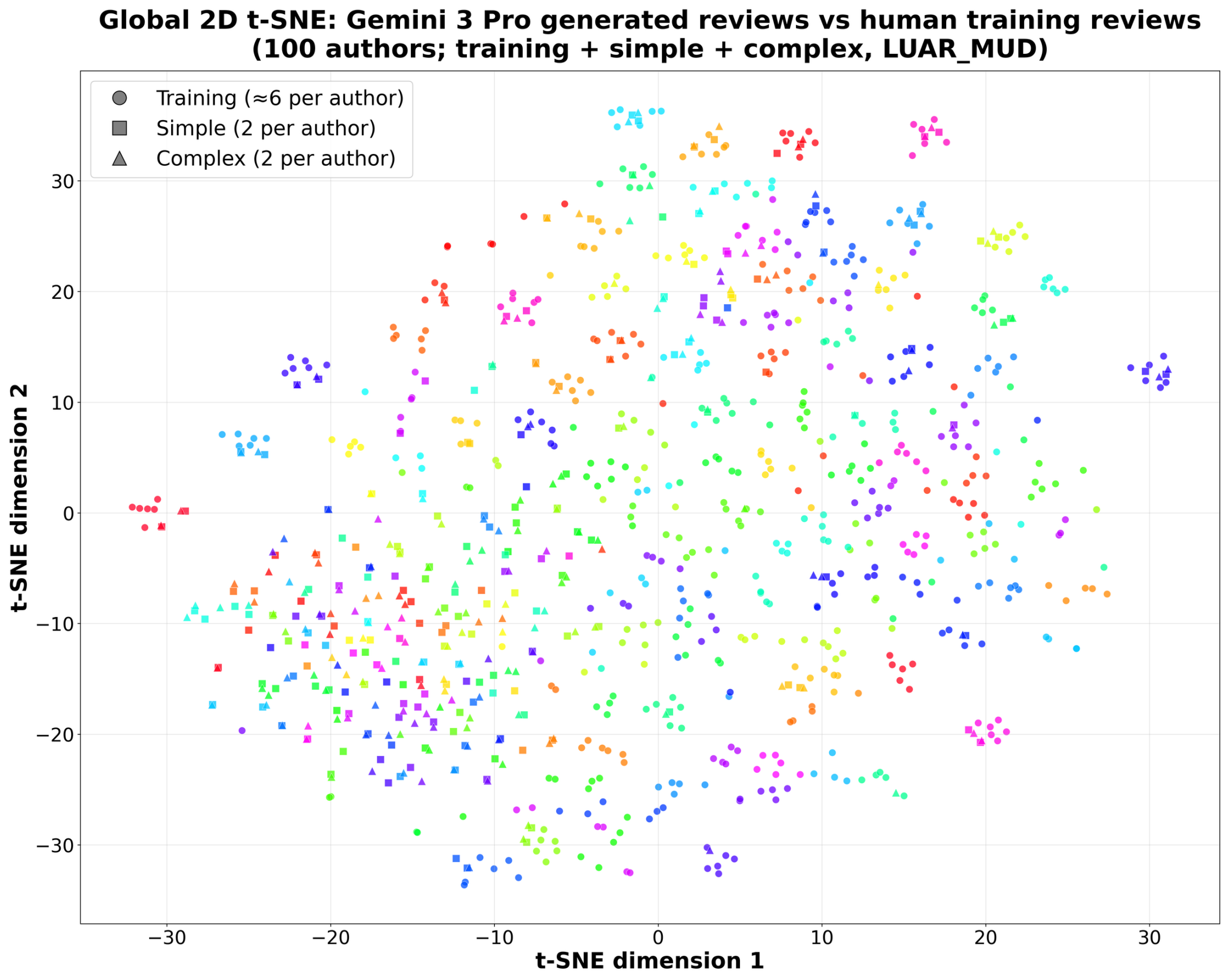

Figure 9: Global two-dimensional *t*-SNE plot of Gemini 3 Pro generated reviews under the simple and complex prompts (two reviews per author per prompt), together with the training reviews (six per author).

Regarding the effect of prompt variant, the global multi-author visualisations showed substantial overlap between reviews generated from the simple and complex prompts, with no clearly perceptible separation in their distributions. This pattern further supports the results reported in Table 5, which indicate that the effect of prompt variant on mimicry performance was generally very small and, for some models, negligible.

Up to this point, Gemini 3 Pro has emerged as the best-performing model in terms of both the distance-to-training metric and the clustering visualisations comparing human-written and machine-generated reviews. However, this apparent advantage requires further scrutiny, particularly with respect to the mechanism underlying its style transfer. Specifically, reduced distances between training and generated reviews may result from the model directly reusing sentences, phrases, or lexical items from the training reviews. Accordingly, an overlap analysis was conducted to assess the extent of textual reuse between the human-written and machine-generated reviews. Substantially elevated overlap was treated as evidence that close clustering or small distances may be confounded by memorisation or direct content reuse, rather than reflecting stylistic transfer alone.

### *5.2 Overlap Analysis*

An overlap analysis was conducted to determine whether high mimicry performance could be attributable to reuse of training text rather than genuine style transfer. The analysis was performed on a per-speaker basis and involved two sets of comparisons (see Figure 10):

1. Overlap between the concatenated reviews in G1 and the concatenated reviews in G2, providing a within-author baseline; and
2. Overlap between each training set and its corresponding generation, namely between the concatenated G1 reviews and the machine-generated review conditioned on G1, and between the concatenated G2 reviews and the machine-generated review conditioned on G2.

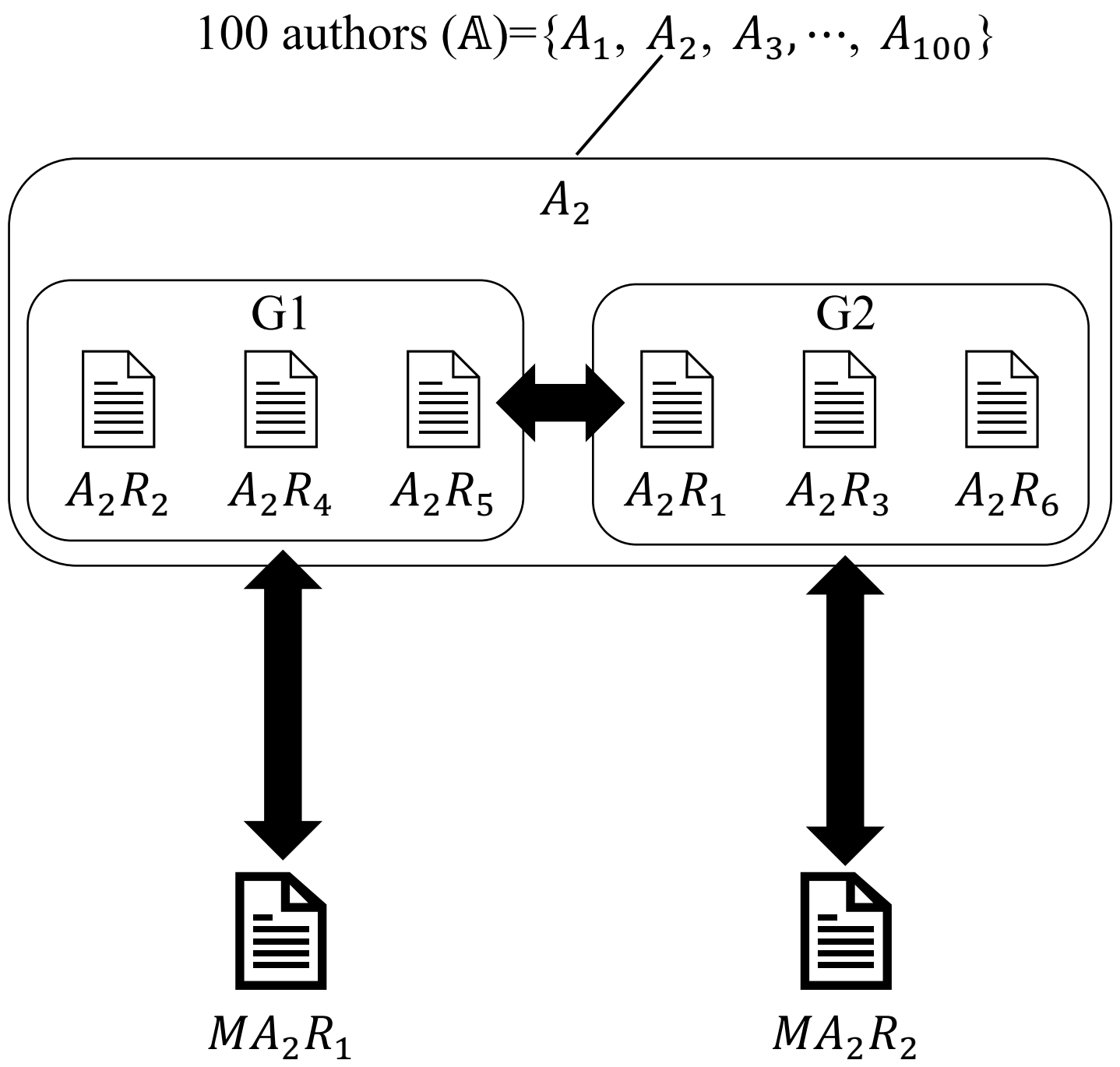

Figure 10: Illustration of the overlap analysis for the second author ($A_2$) using the example given in Figure 5. The horizontal tick arrow = within-author baseline comparison; the vertical thick arrows = comparisons between machine-generated reviews and the corresponding human-written reviews.

The overlap analysis targeted two complementary forms of textual reuse. First, sentence-level character-sequence similarity was computed to detect near-verbatim copying and close paraphrasing. Second, phrase-level reuse was quantified using a trigram-based Jaccard score. The results are summarised in Table 7. In this table, the 100% column reports the number of sentences with exact character-sequence matches, whereas the 85%+ column reports the number of sentences with at least 85% character-sequence similarity. The Jac. > 0.3 column reports the number of sentences for which the trigram-based Jaccard score exceeded 0.3. The Jaccard column reports the mean of the maximum Jaccard scores, averaged across authors.

| LLM | Prompt | 100% | 85%+ | 70–84% | 50–69% | Jac. > 0.3 | Jaccard |
|---|---|---|---|---|---|---|---|
| Human | - | 58 | 55 | 136 | 2470 | 115 | 0.330 |
| Gemini 3 Pro | Simple | 743 | 642 | 1013 | 2314 | 2037 | 0.917 |
| | Complex | 690 | 681 | 1168 | 2500 | 2018 | 0.902 |
| Claude Opus | Simple | 11 | 10 | 22 | 805 | 24 | 0.226 |
| | Complex | 17 | 3 | 37 | 851 | 35 | 0.269 |
| DeepSeek 3.2 | Simple | 34 | 40 | 75 | 869 | 98 | 0.288 |
| | Complex | 47 | 75 | 156 | 1084 | 171 | 0.280 |
| GPT 5.2 | Simple | 2 | 0 | 32 | 686 | 14 | 0.146 |
| | Complex | 3 | 4 | 22 | 665 | 17 | 0.159 |
| GPT 5.2 Pro | Simple | 13 | 14 | 25 | 611 | 35 | 0.294 |
| | Complex | 1 | 0 | 4 | 353 | 4 | 0.095 |
| Grok 4.1 Fast | Simple | 38 | 39 | 89 | 939 | 95 | 0.218 |
| | Complex | 2 | 1 | 5 | 269 | 4 | 0.089 |

Table 7: Summary of the overlap analysis, reporting the number of sentences falling into different bands of character-sequence similarity and the mean trigram-based Jaccard score.

According to Table 7, Gemini 3 Pro exhibited systematic text reuse at a markedly higher level than the other models, with 690–743 exact sentence matches and the average of the individual level max Jaccard scores exceeding 0.9 across both prompt types. In addition, Gemini 3 Pro produced substantially more sentence pairs with high character-sequence similarity than any other model. The following examples illustrate sentences with a Jaccard score of 1, indicating exact matches.

- "I shop on Amazon for everything and think they are great"
- "happy with this purchase I would buy this again"
- "And THAT is all that matters to me"

The following are example sentence pairs with Jaccard scores below 1, indicating partial overlap.

- Jaccard = 0.778
    - Generated: "WHY hadn't I thought of that brand before -- like, maybe, before I had bought the off-brand torture device"
    - Training: "WHY hadn't I thought of that brand before -- like, maybe, before I had bought the Proctor-Silex"
- Jaccard = 0.500:
    - Generated: "Convincing the developers to implement it is the hard part"
    - Training: "Convincing fitbit to implement it is the hard part"
- Jaccard = 0.438
    - Generated: "There are a few things I didn't like about this product that potential buyers should consider"
    - Training: "There are a few things I didn't like about the oven"

By contrast, GPT-5.2 Pro and Grok 4.1 show minimal overlap (1–13 exact matches under the simple prompt and 1–2 under the complex prompt), with mean maximum Jaccard scores (last column) below 0.3, suggesting that their mimicry is less likely to be driven by direct reuse of training material. As a reference point, the human training-text baseline contains 58 exact matches and a mean maximum Jaccard score of 0.330, reflecting natural within-author self-similarity even when reviews target different topics; this indicates that some degree of overlap is expected. Among the remaining models, Claude Opus (11–17 exact matches; Jaccard = 0.226–0.269) and DeepSeek 3.2 (34–47 exact matches; Jaccard = 0.280–0.288) fall into an intermediate range, both below the human baseline. GPT 5.2 shows particularly low overlap across both prompt types (2–3 exact matches; Jaccard = 0.146–0.159). It is notable that, with the exception of Gemini 3 Pro, the LLMs exhibited substantially lower levels of phrase-level reuse and paraphrastic similarity than the human authors. This suggests that human authors rather consistently rely on preferred phrase patterns and recurrent expressions as part of their individual writing style, whereas most LLMs did not reproduce these author-specific patterns to the same extent.

Including explicit anti-copying instructions in the complex prompts (see Appendix A) substantially reduced text reuse for several models. For example, GPT 5.2 Pro decreased from 13 to 1 exact sentence match, with the mean maximum Jaccard score dropping from 0.294 to 0.095, and Grok 4.1 reduced from 38 to 2 exact matches (mean maximum Jaccard: 0.218 to 0.089). These reductions indicate that, for these models, prompting can effectively discourage content copying and promote more faithful style imitation. By contrast, Gemini 3 Pro showed little improvement under the complex prompt, with persistently high overlap. This pattern suggests that its apparent mimicry performance is largely driven by direct reuse of training material rather than transfer of stylistic regularities. In this regard, Claude Opus—ranked second overall—appears more consistent with genuine authorship mimicry, given its relatively low similarity levels and Jaccard scores. The human baseline (mean maximum Jaccard = 0.330) provides a useful reference point: overlap at approximately this level reflects natural within-author self-similarity, whereas Gemini 3 Pro's much higher overlap (mean maximum Jaccard = 0.902–0.917) exceeds plausible human repetition and is more consistent with systematic text reuse than with style learning.

## 5.3 Potential Drivers of Author-Level Variation in Mimicry Performance

The global multi-author visualisations in Figure 7–Figure 9 indicate that mimicry performance varies across authors. For some authors, the generated texts are positioned relatively close to the corresponding human-authored training texts, whereas for others they remain more clearly separated. This suggests that some writing styles are more readily mimicked than others under the present experimental conditions. To investigate possible sources of this author-level variation, this section relates per-author mimicry outcomes to author-level properties derived from the six selected training reviews. Specifically, the following three conceptual predictors, P1–P3, were examined separately:

P1. **Training set consistency**: the extent to which an author's six training documents form a tight cluster.
P2. **Author distinctiveness**: the extent to which an author's style centroid is separated from those of other authors in the evaluated cohort.
P3. **Topic similarity:** the extent to which the generated review aligns with the training reviews in terms of semantic content or topical focus.

### 5.3.1 Training Set Consistency

Training-set-consistency examines whether authors whose training reviews are more internally coherent are also more readily mimicked. Because this construct can be operationalised in more than one way, two complementary metrics, P1M1 and P1M2, were analysed:

- P1M1: **Mean pairwise distance** among the six training reviews.
- P1M2: **Centroid dispersion**, defined as the mean distance of the training reviews to the author centroid.

For each metric, its association with the author's average mimicry distance was assessed using Spearman's rank correlation. Mimicry distance was averaged across the two prompt types, with lower values indicating better mimicry. To account for multiple testing, false discovery rate (FDR) correction was applied across all tests. The results are presented in Table 8.

Table 8: Spearman's rank correlations between training-set-consistency and average mimicry distance for the six LLMs, aggregated across the two prompt types. Results are reported separately for the two consistency metrics, P1M1 and P1M2. FDR-corrected *q* values are shown in parentheses. *** denotes $q < 0.001$.

| LLM | P1M1 | P1M2 |
|---|---|---|
| Gemini 3 Pro | 0.661*** (0.0000) | 0.661*** (0.0000) |
| Claude Opus | 0.512*** (0.0000) | 0.481*** (0.0000) |
| GPT 5.2 Pro | 0.446*** (0.0000) | 0.445*** (0.0000) |
| GPT 5.2 | 0.389*** (0.0002) | 0.397*** (0.0002) |
| DeepSeek 3.2 | 0.230 (0.0538) | 0.241 (0.0538) |
| Grok 4.1 Fast | 0.378*** (0.0005) | 0.365*** (0.0005) |

As shown in Table 8, the training-set consistency is a strong predictor of mimicry performance across all six LLMs. For every model, larger mean pairwise distances (P1M1) and greater centroid dispersion (P1M2) were associated with increased train-generation distances—indicating poorer mimicry performance. In other words, authors whose "reference style" was internally inconsistent were systematically more difficult for models to imitate. This relationship was strongest for Gemini 3 Pro (M1 and M2: $r = 0.661$), remained substantial for Claude Opus 4.5 (P1M1: $r = 0.512$; P1M2: $r = 0.481$) and GPT-5.2 Pro (P1M1: $r = 0.446$; P1M2: $r = 0.445$), and was also clearly present for GPT-5.2 (P1M1: $r = 0.389$; P1M2: $r = 0.397$) and Grok 4.1 Fast (P1M1: $r = 0.378$; P1M2: $r = 0.365$). All of these effects were statistically significant ($q < 0.001$). DeepSeek 3.2 was the only model for which neither metric reached statistical significance.

#### *5.3.2 Author-Distinctiveness*

Author-distinctiveness tests whether more idiosyncratic writing styles are easier for the models to reproduce. Because this construct can be operationalised in more than one way, two complementary metrics, P2M1 and P2M2, were analysed:

- P2M1: **Distance to the nearest author centroid**, defined as a centroid-to-centroid distance.
- P2M2: **Mean distance to the five nearest author centroids**, defined as the average of the five nearest centroid-to-centroid distances.

For each metric, its association with the author's average mimicry distance was assessed using Spearman's rank correlation, following the same procedure used for training-set-consistency. The results are presented in Table 9.

Table 9: Spearman's rank correlations between author-distinctiveness and average mimicry distance for the six LLMs, aggregated across the two prompt types. Results are reported separately for the two distinctiveness metrics, P2M1 and P2M2. FDR-corrected *q* values are shown in parentheses.

| LLM | P2M1 | P2M2 |
|---|---|---|
| Gemini 3 Pro | –0.144 (0.2545) | –0.117 (0.3096) |
| Claude Opus | –0.124 (0.3660) | –0.102 (0.3902) |
| GPT 5.2 Pro | 0.132 (0.2401)) | 0.146 (0.2401) |
| GPT 5.2 | 0.127 (0.2588) | 0.139 (0.2588) |
| DeepSeek 3.2 | 0.045 (0.9867) | –0.002 (0.9867) |
| Grok 4.1 Fast | –0.158 (0.1939) | –0.126 (0.2591) |

In contrast to training-set-consistency, the measures of author-distinctiveness shown in Table 9—operationalised as distance to the nearest author centroid (P2M1) and mean distance to the five nearest author centroids (P2M2)—did not generalise as reliable predictors of mimicry performance. Across the models, the observed associations were very weak and model dependent. Moreover, the direction of the correlations varied by model, alternating between positive and negative values, and none of the relationships reached statistical significance.

### *5.3.3 Topic Similarity*

As a third potential predictor of author-level variation in mimicry performance, topic, or semantic, similarity was examined between the reference reviews used for mimicry and the corresponding generated reviews. Because mimicry performance was evaluated in the LUAR_MUD authorship embedding space, predictors derived from the same representation space could be correlated with mimicry distance partly by construction. To reduce this risk, three training–generation similarity measures, P3M1–P3M3, were computed using independent representation spaces:

- P3M1: **Sentence-BERT**[10] **semantic similarity**, capturing high-level semantic relatedness.
- P3M2: **TF-IDF cosine similarity**, quantifying lexical and topical overlap through shared vocabulary.
- P3M3: **Jaccard similarity** over token sets, measuring direct word-level overlap.

For each author, these measures were calculated by comparing the training reviews with the corresponding generated reviews under both the simple and complex prompting conditions. This allowed topical and lexical similarity to be assessed independently of the LUAR_MUD authorship embedding space used for the main mimicry evaluation.

As with training-set-consistency and author-distinctiveness, the association between each measure and the author's average mimicry distance was assessed using Spearman's rank correlation. The results are presented in Table 10. Negative *r* values indicate that greater topical similarity is associated with lower mimicry distance and, therefore, better distance-based mimicry performance.

[10] Specifically, *all-MiniLM-L6-v2* (https://huggingface.co/sentence-transformers/all-MiniLM-L6-v2) was used with the default settings.

Table 10: Spearman's rank correlations between topic similarity and average mimicry distance for the six LLMs, aggregated across the two prompt types. Results are reported separately for the three distinctiveness metrics, P3M1, P3M2 and P3M3. FDR-corrected q values are shown in parentheses. Asterisks denote statistical significance (*q < 0.05; ***q < 0.001).

| LLM | P3M1 | P3M2 | P3M3 |
|---|---|---|---|
| Gemini 3 Pro | –0.117<br>(0.3078) | –0.368***<br>(0.0004) | –0.405***<br>(0.0001) |
| Claude Opus | –0.045<br>(0.8311) | –0.044<br>(0.8311) | –0.131<br>(0.4868) |
| GPT 5.2 Pro | 0.079<br>(0.7326) | 0.237<br>(0.0593) | 0.081<br>(0.7326) |
| GPT 5.2 | 0.189<br>(0.1977) | 0.073<br>(0.5831) | 0.086<br>(0.5640) |
| DeepSeek 3.2 | 0.005<br>(0.9608) | 0.118<br>(0.3093) | 0.271*<br>(0.0224) |
| Grok 4.1 Fast | 0.055<br>(0.8427) | 0.055<br>(0.8427) | 0.150<br>(0.0522) |

As shown in Table 10, topic and semantic similarity measures exhibited heterogeneous, model-specific relationships with mimicry performance. For some models—notably Gemini 3 Pro—stronger topical alignment between the training reviews and the generated review (P3M2 = –0.368 and P3M3 = –0.405) was associated with better mimicry. This pattern is consistent with the idea that reproducing an author's surface-level stylistic patterns is easier when generation is conditioned on similar topics. However, this finding should be interpreted cautiously in light of Gemini's comparatively high degree of phrase- and sentence-level reuse (see §5.2).

In contrast, other models showed weak or negligible effects of topic similarity. Although the direction of the relationship varied across models—being negative for Claude Opus and positive for others—none of these effects reached statistical significance. DeepSeek 3.2 was the only model to exhibit a statistically significant effect, showing a positive association for P3M3 ($r = 0.271$). Taken together, these results indicate that while topical alignment may modulate mimicry performance for certain models, it does not constitute a uniform or robust predictor across all tested LLMs.

### *5.4 Writing Style Characteristics of LLMs and Human Authors*

In the authorship mimicry tasks and associated analyses, the tested LLMs exhibited distinct style-transfer behaviours, suggesting that each model retains characteristic writing-style tendencies. To examine how similar or different these model-specific styles are from one another, and how they compare with human writing styles, two analyses were conducted:

1. A global distribution analysis of LLM-generated and human-written reviews.
2. Supervised authorship classification across LLMs and human authors.

The results of these analyses are reported separately in §5.4.1 and §5.4.2.

### *5.4.1 Global Distribution Analysis*

First, a global distributional analysis was conducted by projecting both human training reviews and LLM-generated reviews into shared two-dimensional *t*-SNE and UMAP spaces.[11] This analysis was used to examine how similarly or differently the reviews from the seven classes, comprising six LLMs and human authors, were distributed. Because the two visualisations showed highly consistent patterns, the *t*-SNE projection is presented in Figure 11, while the corresponding UMAP plot is provided in Appendix B.

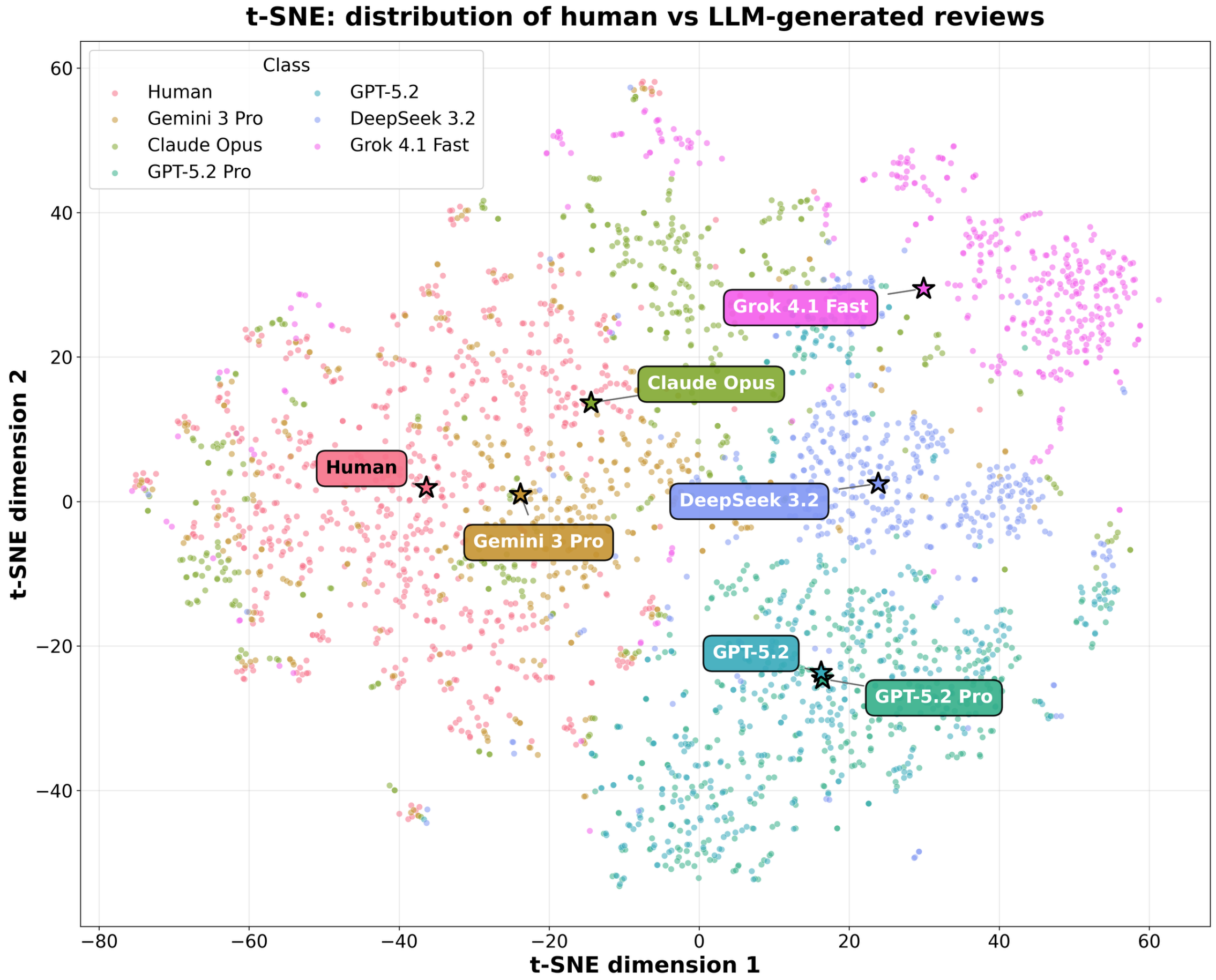

Figure 11: *t*-SNE plot illustrating the distributional positioning of generated reviews from the six LLMs relative to human-written reviews. The centroid of each distribution is marked by a corresponding name label.

As shown in Figure 11, reviews generated by the six LLMs form distinct clusters within the *t*-SNE space, indicating model-specific distributional patterns; i.e., their own unique writing styles. The distributions of GPT 5.2 and GPT 5.2 Pro exhibit substantial overlap, which is unsurprising given their shared architectural lineage. The relative positioning of the LLM clusters with respect to the human-written reviews aligns well with the distance-based mimicry performance rankings (see Table 5). In particular, the centroid of Gemini 3 Pro lies closest to that of the human reviews and their distributions show a fair degree of overlap, followed by Claude Opus. In contrast, the remaining four models—GPT 5.2 Pro, GPT 5.2, DeepSeek 3.2, and Grok 4.1 Fast—are located further away from the human cluster, indicating greater stylistic divergence from the human authors.

[11] The *t*-SNE projection was computed using ‘TSNE’ from ‘sklearn.manifold’, and the UMAP projection was computed using ‘UMAP’ from the ‘umap-learn’ package.

Figure 11 further shows that the generated reviews of Gemini 3 Pro, Claude Opus, GPT 5.2 Pro, and GPT 5.2 are distributed relatively broadly in the *t*-SNE space, to a degree comparable to that observed for human-written reviews. In contrast, the outputs of Grok 4.1 Fast and DeepSeek 3.2 form more compact, tightly clustered distributions. This pattern suggests that the former group of LLMs is capable of producing a wider range of stylistic variation, whereas the latter models exhibit more limited stylistic diversity in their generated reviews.

#### *5.4.2 Supervised Authorship Classification*

To further evaluate whether these model-specific stylistic signatures observed in Figure 11 could be systematically identified, supervised authorship (LLMs and human) classification experiments were conducted. These experiments aimed to assess attribution signals in interpretable feature spaces and to compare their performance against neural embedding-based baselines. Specifically, Random Forest classifiers[12] were trained using bag-of-words (BOW) features (unigrams and bigrams), term frequency-inverse document frequency (TF-IDF)[13] features (unigrams and bigrams) (Manning et al. 2008; Manning and Schütze 2000), and LUAR_MUD embeddings as a baseline representation.

The Random Forest classifier was trained and evaluated using GroupKFold cross-validation, with author identity used as the grouping variable. This ensured that reviews from the same author did not appear in both the training and test sets within the same fold. For the word-based models, feature-importance analyses and SHapley Additive exPlanations (SHAP)[14] explanations (Lundberg and Lee 2017) were subsequently applied to identify the lexical features most indicative of each model's writing style.

The results of the classification experiments are presented in Figure 12 in the form of confusion matrices, shown separately for models trained with BoW features, TF-IDF features, and LUAR_MUD embeddings.

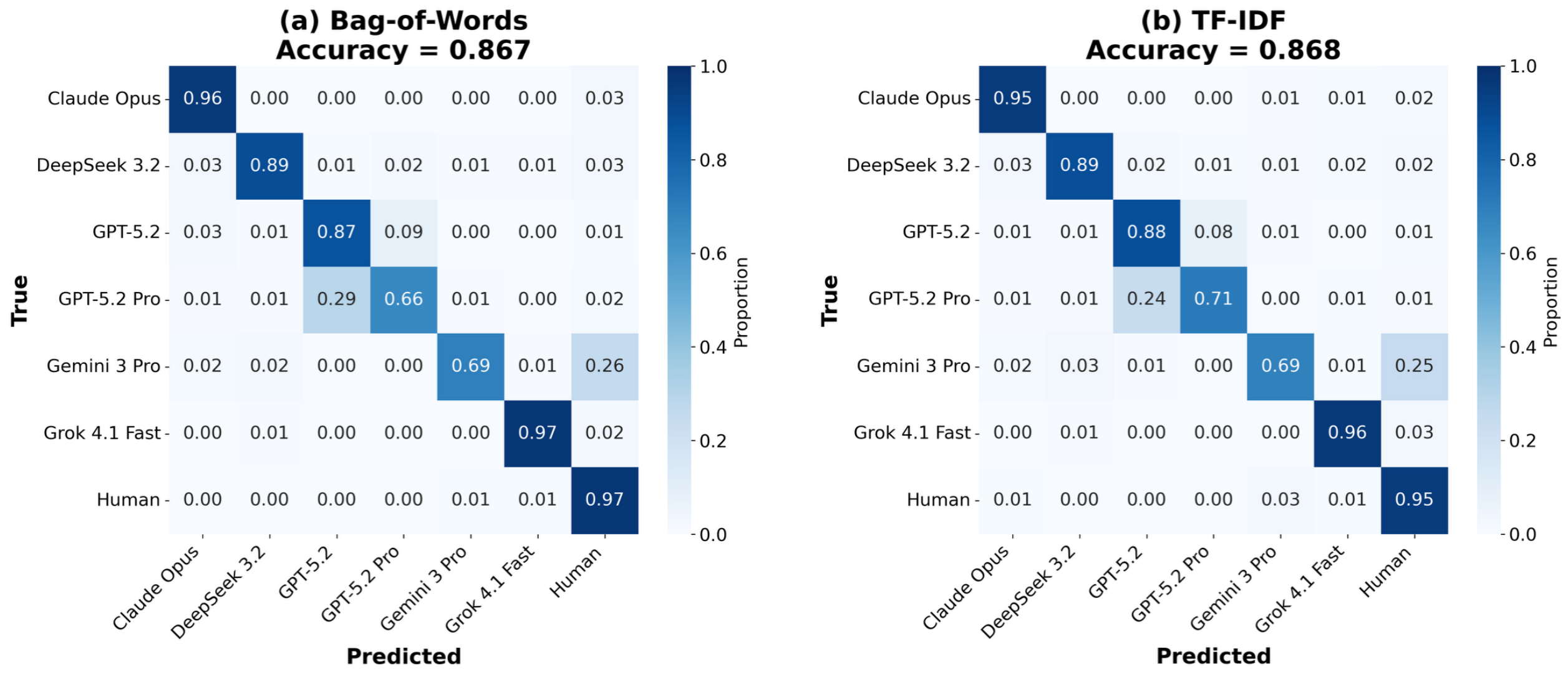

[12] Random Forest classification was performed using the 'RandomForestClassifier' implementation provided by the 'sklearn.ensemble' module in 'scikit-learn'.

[13] To create the BoW and TF-IDF feature representations, a whitespace tokenizer was first applied, after which all punctuation marks except apostrophes were removed from the reviews. This procedure preserved contractions such as "don't", "I've" while removing commas, full stops, and other punctuation marks. BoW features were extracted using 'CountVectorizer', and TF-IDF features were extracted using 'TfidfVectorizer', both from the 'sklearn.feature_extraction.text' module.

[14] SHAP analysis was conducted using the 'shap' Python library.

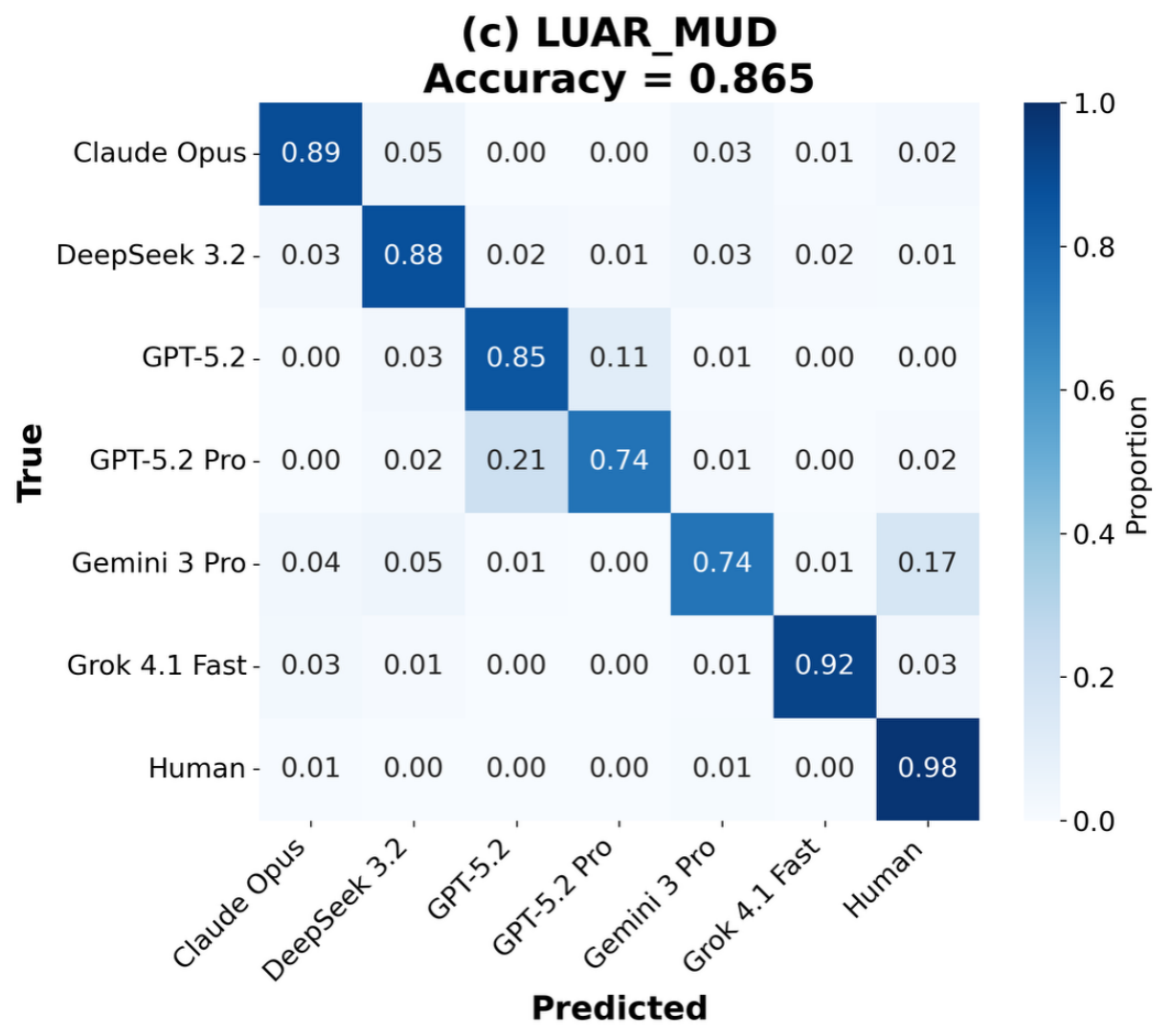

Figure 12: Confusion matrices illustrating classification performance using bag-of-words (BoW) (a), TF–IDF features (b), and LUAR_MUD embeddings (c).

Figure 12 shows that overall classification accuracy is similar across the three representations (BoW: 0.867; TF–IDF: 0.868; LUAR_MUD: 0.865). The confusion patterns among LLM classes are also broadly consistent across the three confusion matrices. The relatively high confusion between GPT 5.2 and GPT 5.2 pro is unsurprising given their substantially overlapping review distributions in Figure 11. Human-written reviews are classified with high accuracy (> 0.95). By contrast, Gemini 3 Pro yields the lowest accuracy (< 0.74), with its reviews most frequently misclassified as human-written, and with additional confusion with Claude Opus and DeepSeek 3.2. This pattern is consistent with Figure 11, where Gemini 3 Pro outputs occupy a central region of the overall distribution and exhibit relatively wide dispersion, making them harder to separate from other sources. Notably, confusion between human-written reviews and Gemini 3 Pro outputs is asymmetric: human reviews are rarely classified as Gemini 3 Pro, whereas Gemini 3 Pro reviews are often classified as human.

Most importantly, the global distributional analysis (Figure 11) and the supervised classification experiment (Figure 12) both indicate that each model exhibits its own characteristic writing style, which is relatively distinct from those of the human authors.

Table 11 lists the 20 most important features (combined unigrams and bigrams) for classifying reviews into the seven classes (six LLMs and human) based on SHAP analyses.

Table 11: The 20 most important features (combined unigrams and bigrams) for BoW and TF-IDF.

| | BoW | | TF-IDF | |
|---|---|---|---|---|
| Rank | Feature | SHAP | Feature | SHAP |
| 1 | I’ve | 0.0260 | I’ve | 0.0240 |
| 2 | it’s | 0.0255 | it’s | 0.0240 |
| 3 | The | 0.0226 | don’t | 0.0204 |
| 4 | don’t | 0.0218 | It’s | 0.0194 |
| 5 | It’s | 0.0192 | I’m | 0.0152 |
| 6 | I’m | 0.0180 | It | 0.0149 |
| 7 | doesn’t | 0.0175 | don’t | 0.0149 |
| 8 | Compared | 0.0164 | you’re | 0.0148 |
| 9 | don’t | 0.0160 | Compared | 0.0145 |
| 10 | you’re | 0.0148 | doesn’t | 0.0144 |
| 11 | It | 0.0139 | The | 0.0133 |
| 12 | I’ve | 0.0131 | I’ve | 0.0124 |
| 13 | you’re | 0.0129 | you’re | 0.0120 |
| 14 | actually | 0.0098 | great | 0.0105 |
| 15 | If | 0.0097 | bought | 0.0098 |
| 16 | Review | 0.0094 | actually | 0.0096 |
| 17 | it’s | 0.0090 | First | 0.0096 |
| 18 | I don’t | 0.0088 | If | 0.0095 |
| 19 | I’m | 0.0084 | I don’t | 0.0092 |
| 20 | It’s | 0.0081 | Review | 0.0090 |

As shown in Table 11, across the two representations (BoW and TF-IDF), the sets of highly weighted features are largely similar, although their rank order differs. Many of the top features are unigrams, and a substantial proportion are common tokens—particularly function words (e.g., “The”) and pronouns (e.g., “I”, “It”)—often appearing in contracted forms (e.g., “I’ve”, “it’s”, “don’t”, “you’re”, “I’m”).

To examine differences in unigram and bigram usage across the six LLMs and human authors, the SHAP analysis was used to identify the 20 most important features (combined unigrams and bigrams) for each of the seven classes. Because the results were highly similar across the two lexical representations, Figure 13 presents the feature importances for the TF-IDF representation only, shown as bar plots.

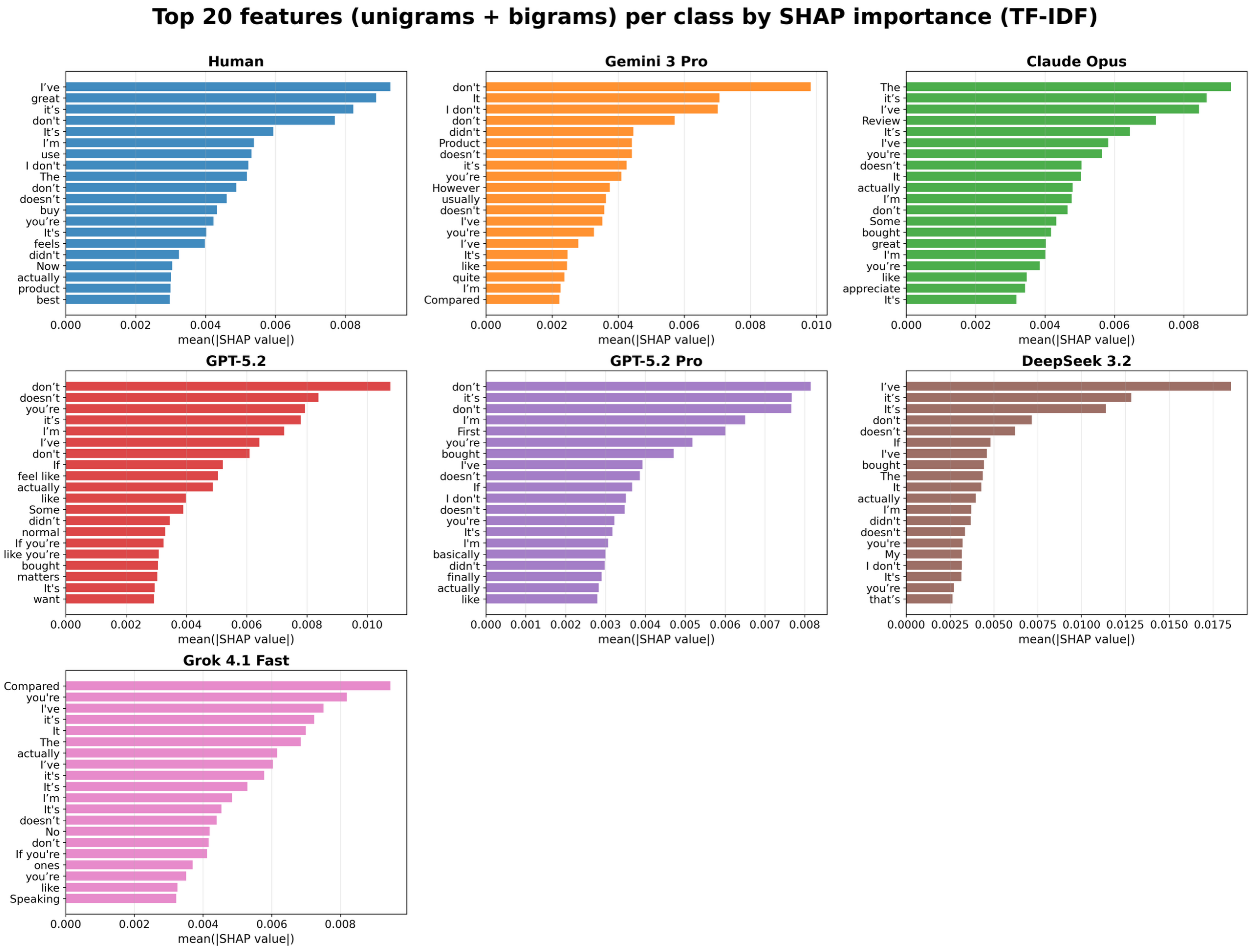

Figure 13: The 20 most important features (combined unigrams and bigrams) for each of the seven classes (six LLMs and human).

Figure 13 shows substantial overlap in the unigrams and bigrams appearing among the top features across the seven classes, foreseeable from the patterns reported in Table 11. This suggests that the classes are distinguished less by the presence or absence of particular features (unigrams or bigrams) than by differences in their relative importance rankings. Nevertheless, a small number of features appear to be comparatively class-specific. For example, “Compared” is the highest-ranked feature for Grok 4.1 Fast, whereas it is ranked much lower for Gemini 3 Pro ($20^{th}$) and does not appear among the top 20 features for the other classes. Similarly, “great” ranks second for the human class but does not appear among the top 20 features for any of the six LLMs, except for Claude Opus 4.5 where it ranks $15^{th}$. Finally, the first-person possessive pronoun “My” appears among the top 20 features only for DeepSeek 3.2.

We note that the SHAP-based feature-importance analysis reported here is preliminary and is included to illustrate the interpretability of the lexical attribution signal rather than as a fully developed result; a complete treatment—identifying which lexical markers are most diagnostic for each author, and by extension most useful for both mimicry-attack and detection strategies—is left to future work.

## 6 Summary of Key Findings

Authorship mimicry was evaluated for six LLMs (Gemini 3 Pro, Claude Opus, GPT 5.2, GPT 5.2 Pro, DeepSeek 3.2, and Grok 4.1 Fast) by providing each model with the same in-context training evidence and asking it to generate a new review in the target author’s style. To isolate stylistic transfer from topic recall, the generation topic was selected to be absent from the training reviews. Each generation context comprised approximately 12 KB of training text (about 2,100 words) drawn from three reviews, and two prompt variants (simple and complex) were tested. The experimental outcomes are summarised below:

- **LLM performance hierarchy:** Under the current experimental setting, the six models differed markedly in authorship mimicry performance. The distance-to-training metric yielded a clear ranking: Gemini 3 Pro > Claude Opus > GPT 5.2 Pro > GPT 5.2 > DeepSeek 3.2 > Grok 4.1.
- **Prompt-type effects:** Prompt-type effects were small, and the complex prompt did not consistently improve performance. Notably, Grok 4.1 Fast performed substantially worse under complex prompting, suggesting that the more elaborate instruction format was not effective for this model under the current experimental conditions.
- **Text reuse and memorisation:** The overlap analysis indicated that Gemini's apparent advantage was largely driven by systematic memorisation and text reuse from the training texts, exceeding the level of repetition observed in human writing, rather than reflecting appropriate stylistic transfer.
- **Predictors of mimicry difficulty:** The factor analysis showed that training-set consistency was the most robust predictor of mimicry difficulty across models: authors whose training reviews formed tighter clusters were consistently easier to mimic. By contrast, author distinctiveness did not generalise as a reliable predictor. Likewise, topic similarity between the training set and the generation topic did not generalise as a reliable predictor, except for Gemini 3 Pro.
- **Corpus-level clustering behaviour:** The global visualisations revealed a consistent macro-level pattern. For most models (GPT 5.2, GPT 5.2 Pro, DeepSeek 3.2, Grok 4.1 Fast), generated texts occupied regions of the embedding space largely separated from the corresponding human training clusters, indicating limited local stylistic convergence at the corpus level. Claude Opus showed more heterogeneous behaviour, with partial alignment for some authors. Gemini 3 Pro exhibited the strongest clustering with training texts.
- **Supervised classification and lexical cues:** In the seven-class attribution task (six LLMs and human), interpretable lexical representations (BoW, TF–IDF) achieved accuracy comparable to neural embeddings (LUAR_MUD). Feature-importance and SHAP analyses further suggested that classes are differentiated less by the presence or absence of specific unigrams or bigrams than by shifts in their relative importance, although a small subset of features appears comparatively class-specific.

## 7 Discussions, Limitations and Future Work

Differences in mimicry performance have been observed among the tested LLMs (Bhandarkar et al. 2024; Chen and Moscholios 2024), with some models outperforming others. Notably, texts generated by lower-performing models (GPT-5.2 Pro, GPT-5.2, DeepSeek 3.2, and Grok 4.1 Fast) form distinct clusters that are clearly separated from human-written texts. This suggests that each model exhibits a characteristic writing style that differs not only from human authors but also from other LLMs. Such model-specific stylistic signatures have been reported in previous studies (O'Sullivan 2025; Zaitsu et al. 2025, 2026), supporting the possibility of detecting AI-generated texts. The presence of distinct clusters separate from human-writing further implies that effective mimicry may involve reproducing at least two types of writing style. The first concerns individual authorship, while the second relates to domain-specific stylistic variation (Biber 1995; Biber et al. 2024). As the task in this study did not explicitly specify domain constraints, this may partly explain the weaker performance of some models. In turn, this suggests that acquiring domain-specific writing conventions prior to attempting authorship mimicry could improve performance for certain LLMs.

The results of the mimicry experiment and subsequent analyses indicate that different LLMs adopted distinct strategies for authorship mimicry. Despite explicit instructions not to copy sentences or phrases, Gemini 3 Pro directly reproduced segments of the training reviews in its outputs. This suggests that, for this model under the present prompting conditions, the practical boundary between style transfer and textual reuse was not always maintained. By contrast, the other LLMs appeared to adhere more closely to the anti-copying instruction. This finding highlights an important methodological principle: embedding-based similarity metrics or authorship verification performance alone are insufficient for evaluating mimicry quality. Such measures should be supplemented with content-overlap analyses to distinguish genuine stylistic imitation from direct copying or near-verbatim reuse.

The prompt design may also have influenced the observed patterns of textual reuse. Because the complex prompt explicitly discouraged the reuse of sentences and phrases, most models produced lower levels of overlap than those typically observed among human-written reviews, where recurrent phraseology and preferred expressions may form part of an author's natural style (Jurafsky et al. 2014). This lower overlap suggests that these models may have successfully internalised the task and reproduced authorship-specific patterns without relying on verbatim reproduction. However, as noted above, compliance with this constraint varied across models: some followed the instruction closely, whereas others only partially respected it (Murthy et al. 2024). Consistent with previous studies (Chen et al. 2025; Sahoo et al. 2024), these findings suggest that prompt effectiveness is not uniform across LLMs. In authorship mimicry tasks, prompt design is therefore a nuanced issue and may require model-specific refinement, particularly when the aim is to encourage stylistic imitation while preventing direct textual reuse (Murthy et al. 2024; Sahoo et al. 2024).

Although authorship impersonation in online forums and blogs has been reported to be challenging for LLMs because these domains are informal and less structurally constrained (Wang et al. 2025), the present study shows that this difficulty may be reduced when a model extensively reuses sentences and phrases from the training texts. As discussed above, however, such reuse should not be interpreted as genuine authorship style transfer, but rather as direct textual copying or near-verbatim reproduction, raising concerns about plagiarism (Lee et al. 2023; McCoy et al. 2023). In such cases, the model may achieve low distance-to-training scores and, for some authors, could potentially evade authorship verification systems, at least under embedding-based evaluation conditions (cf. Zeng and Nini 2026). However, this pattern was not observed uniformly across models, and many LLMs did not achieve comparable levels of impersonation performance.

Importantly, weaker performance against authorship verification systems does not necessarily imply that the generated texts would fail to deceive human recipients. Even texts produced by lower-performing models may be sufficiently plausible to mislead lay readers, particularly in real-world communication settings where recipients may not closely scrutinise stylistic details. This concern is consistent with previous studies showing that humans often have difficulty distinguishing machine-generated texts from human-written texts (Fiedler and Döpke 2025; O'Sullivan 2025; Zaitsu et al. 2025). From a practical misuse perspective, the threshold for successful deception may therefore be considerably lower than the threshold required to evade an authorship verification system.

The factor analysis showed that training-set consistency was the most robust predictor of mimicry difficulty, whereas author distinctiveness did not generalise as a reliable predictor. This asymmetry has practical forensic implications. Authorship verification systems may be more vulnerable to mimicry attacks when the target author writes consistently across documents, regardless of whether that author's style is highly distinctive relative to the broader population. The finding also suggests that authorship obfuscation may be more effective when it deliberately introduces stylistic variability (Bevendorff et al. 2019), rather than merely shifting the text towards more generic or common writing patterns.

The supervised classification experiments showed that interpretable lexical features, namely BoW and TF-IDF, achieved accuracy comparable to that of neural embeddings, specifically LUAR_MUD, in author-class attribution tasks. This finding has practical forensic implications. If relatively simple word-frequency patterns are sufficient to support accurate attribution, then successful mimicry attacks must reproduce not only broader stylistic embedding patterns, but also the specific lexical choices that contribute to attribution decisions. Feature-importance and SHAP analyses, although not fully detailed in the present report, could be extended to identify the lexical markers most diagnostic of each author class. Such analyses would provide useful insight into both potential attack strategies and defensive countermeasures.

Several limitations of the present study should be acknowledged. First, the mimicry experiment was conducted under cooperative prompting conditions: the LLMs were explicitly instructed to imitate the writing style of the target authors. Although this design is appropriate for assessing baseline mimicry potential, it does not fully capture adversarial misuse scenarios in which a motivated attacker actively attempts to evade an authorship verification system. Future work should therefore extend the present framework to adversarial settings. For example, red-teaming experiments could instruct LLMs not only to mimic a target author, but also to minimise detectability by a trained classifier or embedding-based verification system. Iterative evaluation, in which both the generator and detector are refined over successive rounds, would provide a more realistic estimate of the upper-bound risk posed by motivated attackers.

Second, the present study focused on Amazon product reviews, a domain characterised by relatively informal, and weakly structured writing. This provides a useful test case for personal writing-style mimicry, but the findings may not generalise across text types. Future studies should therefore examine authorship mimicry across a wider range of genres and communicative settings, including formal essays, scientific abstracts, social media posts, emails, legal documents, and creative writing. Such work would clarify to what extent LLM-based mimicry is domain-specific or domain-general. Cross-domain experiments would also be valuable; for example, future studies could test whether author-specific evidence from one genre, such as reviews, enables zero-shot or few-shot impersonation in another genre, such as emails. This issue is particularly important for forensic applications, where questioned and known texts often differ in genre, register, or communicative purpose.

Third, the present study was limited to English-language reviews. As a result, it remains unclear whether the observed patterns of mimicry performance, textual reuse, and author-specific susceptibility generalise to other languages. Future research should examine LLM-based authorship mimicry in typologically and sociolinguistically diverse languages, including languages with different writing systems, morphological structures, and conventions of written style.

Finally, this study evaluated six LLMs over a single time period (December 2025–January 2026). Given the rapid pace of development in generative AI systems, these findings should be interpreted as a time-specific snapshot rather than a stable estimate of long-term capability. As emphasised by Thompson and Ishihara (2025: 23) and Ishihara (2026: 11), future longitudinal research is therefore needed to track how authorship mimicry evolves across successive model generations. Such work would help determine whether performance improves, plateaus, or declines over time, and whether emerging capability shifts introduce new risks for existing authorship verification systems. Regular reassessment will thus be essential to maintaining the forensic relevance of both detection methods and evaluation benchmarks.

## Appendix A. Prompt Variants

Table 12: Prompt types: Simple and Complex. The selected generation topic is inserted at {generation_topic}.

| | |
|---|---|
| Simple | *Below are example Amazon product reviews written by the same person.*<br>*Learn their writing style and write a new Amazon product review about a product in the category: {generation_topic}.*<br><br>*Constraints:*<br>*Write around 600–900 words.*<br><br>*==== EXAMPLE REVIEWS START ===*<br>⋮<br>⋮<br>*==== EXAMPLE REVIEWS END ====*<br><br>*Now write the new review.* |
| Complex | *You are an expert at mimicking writing style while changing the topic.*<br>*Below are several example Amazon product reviews written by the SAME anonymous author.*<br>*Your task is to write a NEW product review in the same writing style (sentence rhythm, vocabulary, level of detail, tone), but about a product in the following category: {generation_topic}.*<br><br>*Constraints:*<br>*1. Do not copy any specific sentences or phrases from the examples.*<br>*2. Write around 600–900 words.*<br>*3. Write a single coherent review with natural structure.*<br>*4. Do not mention that you are an AI model or that you are imitating anyone.*<br><br>*==== EXAMPLE REVIEWS START ====*<br>⋮<br>⋮<br>*==== EXAMPLE REVIEWS END ====*<br><br>*Now write the new review.* |

## Appendix B. UMAP plot

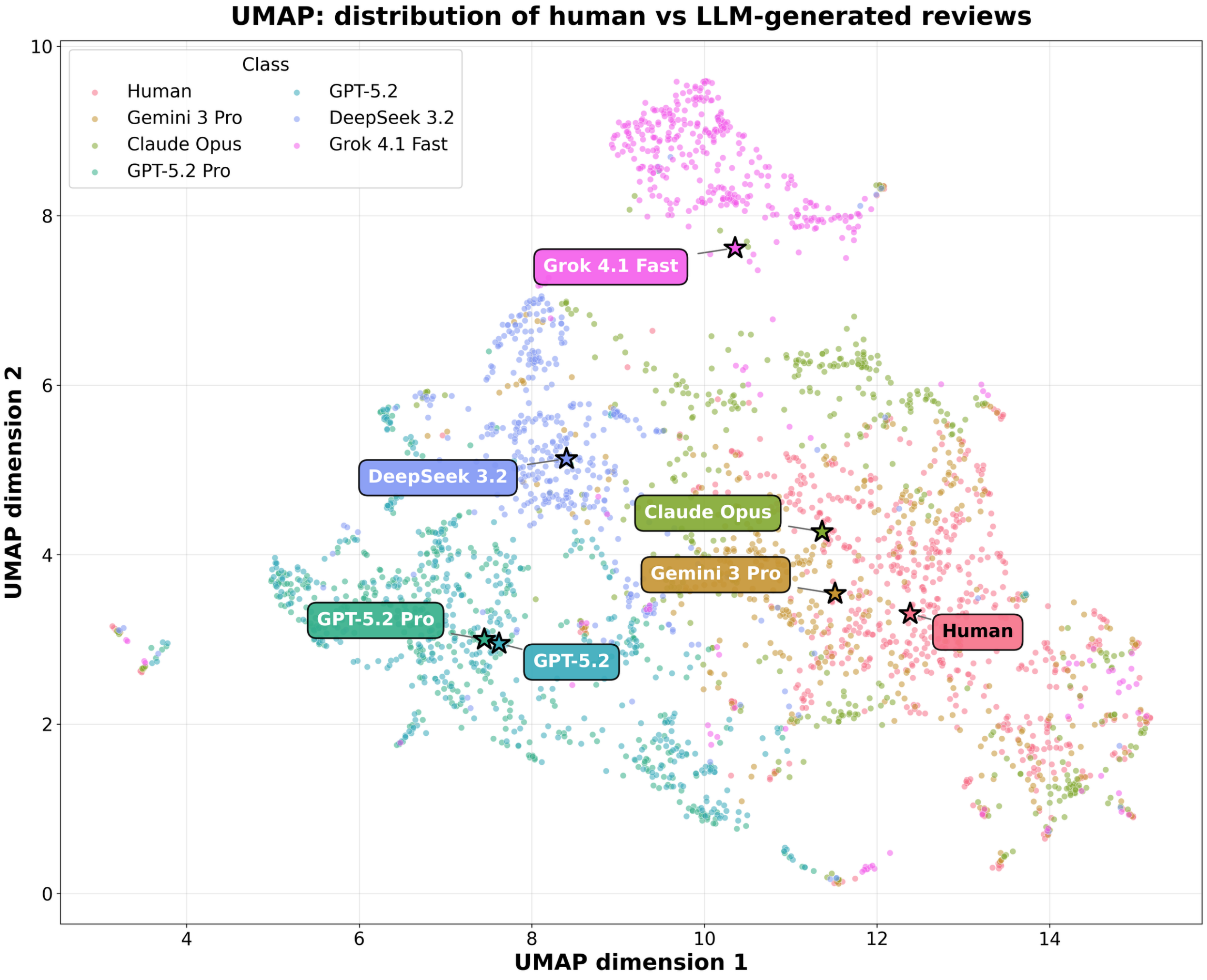

Figure 14: UMAP plot illustrating the distributional positioning of generated reviews from the six LLMs relative to human-written reviews. The centroid of each distribution is marked by a corresponding name label.